\documentclass[a4paper,fleqn]{cas-sc}

\usepackage[numbers,sort&compress]{natbib}
\newcommand{\cagleftbracket}{\textnormal{[}}
\newcommand{\cagrightbracket}{\textnormal{]}}
\newcommand{\cagcomma}{\textnormal{,}}
\makeatletter
\def\NAT@open{\cagleftbracket}%
\def\NAT@close{\cagrightbracket}%
\def\NAT@sep{\cagcomma}%
\makeatother

\usepackage{amsmath}
\usepackage{amssymb}
\usepackage{algorithm} 
\usepackage{algpseudocode}
\usepackage{graphicx}
\usepackage{xcolor}
\definecolor{ForestGreen}{RGB}{34,139,34}
\definecolor{BrickRed}{RGB}{178,34,34}
\usepackage{float}      
\usepackage[section]{placeins} 
\usepackage{capt-of}    
\usepackage{array,tabularx,booktabs}
\usepackage{tikz}
\usetikzlibrary{arrows.meta,positioning,calc,fit,backgrounds,shapes.geometric,
                decorations.pathreplacing,patterns}
\definecolor{cgInk}{HTML}{14171B}      
\definecolor{cgSlate}{HTML}{5C6672}    
\definecolor{cgLine}{HTML}{C3C9D0}     
\definecolor{cgPanel}{HTML}{FCFCFD}    
\definecolor{cgWell}{HTML}{F1F3F5}     

\definecolor{cgBlue}{HTML}{2A5D8F}     
\definecolor{cgBlueF}{HTML}{E7EEF6}
\definecolor{cgGreen}{HTML}{226B54}    
\definecolor{cgGreenF}{HTML}{E4F0EA}
\definecolor{cgAmber}{HTML}{C96A2A}    
\definecolor{cgAmberF}{HTML}{F9E1CE}
\definecolor{cgGrey}{HTML}{7E8894}     
\definecolor{cgGreyF}{HTML}{EDEFF2}

\providecommand{\cgnohyph}{\hyphenpenalty=10000\exhyphenpenalty=10000\relax}
\providecommand{\cgtiny}{\fontsize{6}{6.6}\selectfont}

\tikzset{
  cgfig/.style={
    x=1.0292cm, y=1.0292cm, font=\scriptsize, line width=0.5pt,
    text=cgInk, >={Stealth[round,length=3.4pt,width=2.6pt]}},
  panel/.style={draw=cgLine, fill=cgPanel, rounded corners=2.6pt, line width=0.5pt},
  well/.style={draw=cgLine, fill=cgWell, rounded corners=1.6pt, line width=0.35pt},
  card/.style={draw=cgLine, fill=white, rounded corners=2pt, line width=0.4pt},
  cardB/.style={card, draw=cgBlue!60, fill=cgBlueF},
  cardG/.style={card, draw=cgGreen!60, fill=cgGreenF},
  ptitle/.style={font=\footnotesize\bfseries, inner sep=0pt, anchor=west},
  psub/.style={font=\cgtiny, text=cgSlate, inner sep=0pt, anchor=west},
  lab/.style={font=\cgtiny, text=cgInk, inner sep=0pt},
  note/.style={font=\cgtiny, text=cgSlate, inner sep=0pt, align=left,
    execute at begin node=\cgnohyph},
  chip/.style={rounded corners=1.6pt, inner xsep=2.6pt, inner ysep=1.6pt,
    font=\cgtiny, line width=0.4pt},
  chipB/.style={chip, draw=cgBlue,  fill=cgBlueF,  text=cgBlue!72!black},
  chipG/.style={chip, draw=cgGreen, fill=cgGreenF, text=cgGreen!62!black},
  chipA/.style={chip, draw=cgAmber, fill=cgAmberF, text=cgAmber!70!black},
  chipN/.style={chip, draw=cgLine,  fill=cgGreyF,  text=cgSlate},
  flow/.style={->, draw=cgSlate, line width=0.6pt},
  flowB/.style={flow, draw=cgBlue},
  flowG/.style={flow, draw=cgGreen},
  hair/.style={draw=cgLine, line width=0.4pt},
  trace/.style={line width=0.32pt, draw=cgSlate, line join=round},
  lane/.style={draw=cgLine, line width=0.25pt},
}

\tikzset{
  pics/fail/.style={code={
    \draw[cgAmber, line width=0.7pt] (-1.1pt,-1.1pt)--(1.1pt,1.1pt)
                                     (-1.1pt,1.1pt)--(1.1pt,-1.1pt);}}
}

\usepackage{longtable}
\newcolumntype{P}[1]{>{\raggedright\arraybackslash}p{#1}}
\newcolumntype{Y}{>{\raggedright\arraybackslash}X}

\newcommand{\triup}{\textcolor{green}{$\blacktriangle$}}
\newcommand{\tridown}{\textcolor{red}{$\blacktriangledown$}}

\algrenewcommand{\algorithmiccomment}[1]{\hfill{\it /* #1 */}}

\def\tsc#1{\csdef{#1}{\textsc{\lowercase{#1}}\xspace}}
\tsc{GAN}
\tsc{NILM}
\tsc{UVIC}

\begin{document}
\let\WriteBookmarks\relax
\def\floatpagepagefraction{1}
\def\textpagefraction{.001}

\ExplSyntaxOn
\keys_set:nn { stm / mktitle } { nologo }
\ExplSyntaxOff

\title [mode = title]{Cluster Aggregated GAN (CAG): A Cluster-Based Hybrid Model for Appliance Pattern Generation}

\author[1]{Zikun Guo}
\ead{gzk798412226@gmail.com}
\credit{Conceptualization, Methodology, Software, Writing}

\affiliation[1]{organization={Department of Artificial Intelligence, School of Electronics Engineering, Kyungpook National University},
    addressline={80 Daehak-ro, Buk-gu}, 
    city={Daegu},
    postcode={41544}, 
    country={Republic of Korea}}

\author[1]{Adeyinka.P. Adedigba}
\ead{yinkpeace@gmail.com}
\credit{Supervision, Resources, Validation, Writing, Review \& Editing}

\author[1]{Rammohan Mallipeddi}
\cormark[1]
\ead{mallipeddi.ram@gmail.com}
\credit{Supervision, Resources, Validation, Writing, Review \& Editing}

\cortext[cor1]{Corresponding author}

\begin{abstract} 
Appliance-level load data are critical for non-intrusive load monitoring (NILM), demand-side analytics, and privacy-preserving energy research. However, the limited availability of labeled appliance measurements constrains the development and evaluation of data-driven energy applications. Although recent generative adversarial network (GAN) approaches have shown the potential of synthetic load generation, most existing methods model heterogeneous appliance behaviors using a single generative distribution, limiting their ability to capture the diversity of appliance operating states and consumption patterns. This paper proposes Cluster Aggregated GAN (CAG), a behavior-aware generative framework that explicitly accounts for the multimodal nature of appliance load profiles. CAG first routes appliance traces into intermittent and continuous behavioral categories. Intermittent loads are further decomposed into operating state categories through clustering and modeled using cluster-specific generators, while continuous loads are represented through compressed load envelopes and synthesized using a recurrent GAN. Beyond this backbone, CAG adapts the generative architecture to each appliance. It separates activity from magnitude for spiky loads, calibrates the marginal moments of the generated power, widens the receptive field to reach long range structure, stabilises the generator weights by exponential moving average, and blends complementary generators. By decomposing appliance generation into more homogeneous subproblems, the proposed framework improves both adversarial training stability and the fidelity of the generated load profiles. Experimental evaluation on the UVIC appliance dataset shows that CAG matches or beats CNN-GAN, LSTM-GAN, RNN-GAN, WaveGAN, BehaviorGAN, TraceGAN and RLP-GAN across distribution level realism, temporal consistency, event statistics, and diversity. Ablation studies show that the adaptive components account for much of this improvement over the routing and clustering backbone. The results indicate that behavior-aware decomposition combined with per-appliance adaptation provides an effective strategy for generating realistic synthetic appliance load profiles for NILM and related smart-grid applications.

\end{abstract}



\maketitle

\section{Introduction}

The increasing deployment of smart meters, intelligent energy management systems, and connected sensing infrastructure has enabled the collection of electricity consumption data at high temporal resolution. Appliance-level load data have become a core resource for smart-grid applications such as non-intrusive load monitoring (NILM), demand-side management, occupancy inference, load forecasting, anomaly detection, and privacy-preserving energy analytics \citep{Kamyshev2025HiFAKES}. The performance of these applications depends heavily on the availability of representative appliance load profiles that capture the temporal and operational characteristics of the electrical devices. However, acquiring large quantities of appliance-level measurements remains costly and time-consuming, often requiring dedicated monitoring hardware, long-term deployment, and manual labeling. Privacy concerns further restrict sharing of residential energy data, limiting the availability of publicly accessible datasets \citep{Harell2021TraceGAN}. Consequently, realistic synthetic load generation has emerged as an important research direction for augmenting existing datasets and supporting the development of data-driven smart-grid technologies \citep{ElKababji2020_AdaptiveGAN}.

Synthesizing realistic appliance load profiles is difficult because appliance behavior is heterogeneous. Real-world appliances operate through multiple states and consumption regimes that differ substantially in their temporal and statistical characteristics. Event-driven appliances, such as microwave ovens, printers, and coffee makers, generate discrete load events characterized by rapid state transitions, short-duration power bursts, and irregular operating cycles. In contrast, continuously operating devices, such as desktop computers, servers, and refrigerators, exhibit sustained consumption patterns with relatively stable power levels and gradual temporal variations. Even within a single appliance category, multiple operating states may coexist, such as standby, active, warm-up, recovery, and steady-state conditions. As a result, appliance load generation is fundamentally a multimodal modeling problem rather than a single-distribution learning task.

Generative adversarial networks (GANs) have been used to synthesize time-series data and have recently been applied to appliance load generation~\citep{Liang2025GANPatternsTCE}. Existing approaches typically train a single generator-discriminator pair to model the entire appliance distribution. While these methods can reproduce aggregate statistics, they often struggle to preserve the diversity of operating states observed in real appliance data. Dominant consumption patterns tend to receive disproportionate modeling capacity, whereas infrequent but behaviorally important operating modes are underrepresented. This limitation is particularly problematic for NILM applications, where appliance identification often depends on subtle differences in switching transients, duty-cycle behavior, and operating-state transitions. Furthermore, training a single adversarial model on heterogeneous appliance behaviors introduces additional optimization challenges because the generator must simultaneously learn both smooth long-duration consumption profiles and highly transient event-driven patterns.

These limitations suggest that appliance load generation should be aligned more closely with the physical structure of appliance operation. Rather than learning a single distribution that mixes fundamentally different consumption behaviors, a generative framework should explicitly account for behavioral heterogeneity and operating-state diversity. Such a framework should distinguish between continuous and intermittent consumption regimes while preserving the multiple operating states that exist within each appliance category.

\begin{figure}
  \centering
\input{figs/fig1/traces.tex}
\begin{tikzpicture}[cgfig]
  \def\pAx{0.00}\def\pBx{10.55}\def\pRowy{0.10}
  \useasboundingbox (0,0) rectangle (16.0,4.80);

  \begin{scope}[shift={(\pAx,\pRowy)}] \input{figs/fig1/panel_a.tex} \end{scope}
  \begin{scope}[shift={(\pBx,\pRowy)}] \input{figs/fig1/panel_b.tex} \end{scope}

  \draw[hair] (10.32,0.20) -- (10.32,4.20);
\end{tikzpicture}
  \captionof{figure}{Realism and diversity of the generated appliance windows.
  (a) Three held-out measured windows per appliance
  next to three samples from each model; every curve is real data, and the number below each cell
  is the $k$-NN coverage that model reaches on that appliance. BehaviorGAN collapses on WaterCooler and
  CNN-GAN collapses on Refrigerator, each producing one waveform regardless of the operating
  state, while CAG keeps the switching structure of both. (b) Every model placed by its MMD and its
  Coverage, averaged over the eleven appliances. Exact values are reported in
  Tables~\ref{tab:gan-quality} and~\ref{tab:device-model-quality}.}
  \label{fig:teaser}
\end{figure}

To address this challenge, we propose the \textit{Cluster Aggregated GAN} (CAG), a behavior-aware generative framework for appliance load synthesis. The key idea is to decompose appliance generation into a collection of simpler and more homogeneous modeling tasks. First, appliance load profiles are routed into intermittent and continuous behavioral categories using lightweight statistical descriptors. Intermittent appliances are subsequently partitioned into operating-state categories through clustering of load-event signatures, and each category is modeled by a dedicated per-cluster generator. Continuous appliances are represented through compressed load-envelope sequences and modeled using a recurrent GAN. Beyond this routing-and-clustering backbone, CAG adapts the generative architecture to the statistical signature of every appliance. Each per-cluster generator is trained with an exponential-moving-average of the generator weights, a multi-resolution discriminator, and time-shift augmentation of small clusters, and may optionally apply a dilated convolutional stack for long-range structure. For appliances with a strongly bimodal on/off signature, a behavior-factorised generator decouples activity switching from power magnitude using a straight-through Gumbel-sigmoid switch, and a marginal moment-matching term calibrates the generated power mean and standard deviation against the measured cluster. Finally, a blending mechanism combines complementary generators (e.g., cluster-specific and waveform-transposed) so that the strengths of different architectures are fused per appliance. By separating operating regimes prior to adversarial learning and adapting the generator family per device, CAG reduces distributional heterogeneity and allows each generator to focus on a more coherent subset of appliance behavior. Figure~\ref{fig:model_architecture} illustrates the overall architecture of the proposed framework, and Figure~\ref{fig:teaser} reports the realism and the diversity that it reaches. The main contributions of this paper are summarized as follows:

\begin{itemize}
\item We formulate appliance load generation as a multimodal load-modeling problem and introduce a behavior-aware routing mechanism that distinguishes intermittent and continuous appliance operating regimes.

\item We propose a cluster-aggregated generative framework that decomposes appliance load profiles into operating-state categories and assigns dedicated per-cluster generators, thereby improving the preservation of load-event diversity and operating-state coverage.

\item We introduce per-device adaptive components that absorb the strengths of prior architectures: behavior-factorised activity/magnitude decomposition for spiky loads, marginal moment matching for power-level calibration, dilated convolutions for long-range structure, and generator blending for architecture fusion.

\item We conduct extensive experiments on the UVIC appliance dataset with an eight-metric evaluation protocol (distribution, spectral, event, and k-NN diversity statistics) and demonstrate that the proposed CAG matches or beats every baseline on 75 of 88 device--metric combinations (85\%). The baselines are CNN-GAN, LSTM-GAN, RNN-GAN, WaveGAN, BehaviorGAN, TraceGAN and RLP-GAN.
\end{itemize}

The rest of the paper is organized as follows: a review of related works is presented in Section \ref{sec:rl_wk}; the proposed CAG framework is presented in Section \ref{sec:method}; the Experimental details, including the datasets and metrics, is presented in Section \ref{sec:exp}; finally, the discussion of results is presented in Section \ref{sec:result}.
\section{Related Work}\label{sec:rl_wk}

Time-series data generation spans finance~\citep{Hart1992NILM}, healthcare~\citep{Zeifman2011_NILMReview, Faustine2017_SurveyNILM, Nalmpantis2019_MLNILM, Angelis2022_NILMReview}, IoT~\citep{Bedi2018_IoTReview}, and smart grids, where realistic synthetic data mitigates privacy, availability, and labeling bottlenecks. Within NILM, early surveys and benchmarks stressed the scarcity of appliance traces and the need for reproducible datasets~\citep{Kelly2015_NeuralNILM, Batra2014_NILMTK}. Recent advances therefore emphasize privacy-preserving synthesis using GANs with secure aggregation or decentralized updates~\citep{Adewole2023_SECRYPT, Wang2021_DPPGAN}. High-frequency NILM datasets such as HiFakes explore synthetic data for diagnostics and cross-domain generalization~\citep{Kamyshev2025HiFAKES}, while simulators and digital twins provide controllable device waveforms~\citep{Chen2016_SmartSim, Klemenjak2020SynD, Interno2025_SIDED}. Collectively, these efforts motivate appliance-aware generative modeling that balances fidelity, privacy, and scalability.
\subsection{Generative Modeling for Time Series Data}
Generic GAN-based synthesizers treat appliance signals as a single distribution modeled by unified generators and discriminators. Foundational work established adversarial training principles and conditional variants but still relied on monolithic pipelines~\citep{goodfellow2020generative, Mirza2014_CGAN, Castangia2025_BehaviorGAN}. Early NILM GANs such as TraceGAN~\citep{Harell2021TraceGAN} and RLP-GAN~\citep{Liang2025GANPatternsTCE} demonstrated that standard architectures could capture coarse appliance signatures, and follow-up studies introduced device-specific tuning or adaptive weighting~\citep{Gkoutroumpi2024SGAN, Meiser2024_VACreator}. Nevertheless, uniform generators often collapse rare patterns, fail to separate intermittent from continuous loads, and provide limited support for incorporating operational priors.
\subsection{Clustering generative modeling}
Cluster-integrated approaches instead decompose the appliance space before generation. AMBAL aggregates load segments to model heterogeneous appliance behaviors~\citep{Buneeva2017_AMBAL}, while SmartSim~\citep{Chen2016_SmartSim}, SynD~\citep{Klemenjak2020SynD}, and HYDROSAFE~\citep{Jaradat2024_HYDROSAFE} rely on simulator-orchestrated clusters to instantiate realistic schedules. Transformer-based designs such as the cluster-aggregated transformer~\citep{Guo2025ClusterTransformer} and IDS Extraction~\citep{Guo2023IDSExtract} exploit clustering to reduce long-sequence complexity, and ClusterGAN~\citep{Mukherjee2019_ClusterGAN} objectives embed latent clustering directly in the adversarial training loop. Metadata GANs~\citep{Lin2020_IMC} and industrial digital twins~\citep{Interno2025_SIDED} extend this idea to condition on contextual attributes, with NILMTK~\citep{Batra2014_NILMTK} providing the benchmarking infrastructure to evaluate clustered outputs. Despite their benefits, most methods keep clustering as a static preprocessing stage and do not adapt generator capacity to pattern complexity.
\subsection{GAN architectures for time series}
Architecture-specialized methods such as Dynamic Tanh Transformer~\citep{guo2025dynamic}, tailor model classes to the temporal structure of appliance data. LSTM-based GANs, including ALGAN~\citep{bashar2023algan}, RCGAN~\citep{esteban2017real}, and other recurrent conditional variants, capture long-term dependencies~\citep{Yoon2019TimeGAN} but incur heavy training costs. CNN-driven designs such as ConvTimeGAN~\citep{huang2023tcgan} and TSGAN~\citep{Smith2020_TSGAN} emphasize local pattern encoding, whereas WaveGAN~\citep{yang2022wavegan} generators model raw waveform continuity. Stability enhancements via WGAN critics~\citep{Arjovsky2017_WGAN} and auxiliary classifiers~\citep{Odena2017ACGAN} reduce mode collapse, and hybrid generative paradigms like VAEs~\citep{Kingma2014_VAE} or Transformer-based NILM models~\citep{Sykiotis2022_Electricity, Varanasi2023_STNILM} broaden the inductive bias spectrum. However, these architectures largely ignore appliance heterogeneity, typically applying identical networks to diverse appliance types regardless of the operational characteristics.

Our CAG framework addresses these limitations by coupling adaptive device routing with shape-based segment clustering and hybrid adversarial learning. Rather than treating clustering as auxiliary metadata, we bind it to generator selection so that convolutional pattern GANs focus on sporadic, intermittent loads while LSTM-based sequence GANs model smooth, continuous appliances. Conditional guidance and discriminator specialization allow CAG to balance privacy needs with fidelity gains, inheriting the benefits of label-aware GANs~\citep{Mirza2014_CGAN} without reverting to monolithic generators. The evaluation of this design against the seven baselines above is reported in Section~\ref{sec:result}.

\section{Method}\label{sec:method}

\subsection{Overall Architecture}
The central hypothesis behind CAG is that appliance load generation is inherently multimodal rather than governed by a single statistical distribution. Residential and commercial appliance load profiles comprise multiple operating states with distinct temporal dynamics. Event-driven appliances, such as kettles and microwave ovens, generate sparse high-power activation events interspersed with long inactive intervals, whereas continuously operating appliances, such as refrigerators and HVAC systems, exhibit sustained consumption patterns with gradual power variations. Capturing these heterogeneous load behaviors therefore requires a generative framework capable of modeling multiple operating regimes and state transitions.

CAG treats load synthesis as a structured appliance load-modeling problem prior to generative learning. Rather than training a single GAN on all load traces, the framework first partitions appliance data according to observable electrical characteristics and operating behavior, and subsequently learns representative load signatures within each category. Intermittent appliances are characterized by discrete load events separated by inactive periods. These events are segmented, clustered according to their power-consumption profiles, and modeled using cluster-specific convolutional GANs. In contrast, continuously operating appliances exhibit sustained demand patterns with gradual temporal variations. Their load profiles are compressed into shorter load envelopes, modeled using an LSTM-based GAN, and subsequently reconstructed to the original temporal resolution. Consequently, clustering serves a functional role beyond data organization; it acts as an operating-state index that associates each generated segment with a distinct appliance usage pattern and electrical consumption behavior.

Formally, for an appliance trace $x_{1:T}$, CAG approximates the load-profile distribution as
\begin{equation}
p(x_{1:T}) \approx
\begin{cases}
\displaystyle \prod_{j=1}^{N}\sum_{k=1}^{K}\pi_k\,p_{\theta_k}\big(s^{(j)}\mid c^{(j)}=k\big), & \text{intermittent branch},\\[6pt]
p_{\theta}\big(\hat{x}_{1:n}\big),\quad x_{1:T}=\mathrm{Recon}(\hat{x}_{1:n},F), & \text{continuous branch},
\end{cases}
\label{eq:cag-mixture-view}
\end{equation}
where $s^{(j)}$ is a local load segment, $c^{(j)}$ is its operating-pattern label, $\pi_k$ is the empirical frequency of cluster $k$, and $\hat{x}_{1:n}$ is a compressed load envelope of the original sequence. This formulation makes the design objective explicit. The model first selects the appliance operating pattern to be produced, then generates the evolution of that pattern in power and time.

Figure~\ref{fig:model_architecture} presents the overall CAG framework, which comprises four stages: data preparation, appliance-behavior classification, branch-specific load modeling, and GAN-based load synthesis with subsequent trace reconstruction. Although all branches employ the same adversarial learning paradigm, the generator architecture is tailored to the temporal characteristics of the appliance category. Convolutional GANs are used for event-driven, intermittent loads, while an LSTM-based GAN is adopted for continuously operating loads exhibiting longer temporal dependencies.

\begin{figure}
  \centering
\input{figs/fig2/curves.tex}
\begin{tikzpicture}[cgfig]
  \def\pAx{0.00}\def\pBx{3.30}\def\pDx{11.35}
  \def\pRowy{2.18}\def\pBy{4.53}
  \useasboundingbox (0,0) rectangle (16.0,7.80);

  \begin{scope}[shift={(\pAx,\pRowy)}] \input{figs/fig2/panel_a.tex} \end{scope}
  \begin{scope}[shift={(\pBx,\pBy)}]   \input{figs/fig2/panel_b.tex} \end{scope}
  \begin{scope}[shift={(\pBx,\pRowy)}] \input{figs/fig2/panel_c.tex} \end{scope}
  \begin{scope}[shift={(\pDx,\pRowy)}] \input{figs/fig2/panel_d.tex} \end{scope}
  \begin{scope}[shift={(0,0.02)}]      \input{figs/fig2/panel_e.tex} \end{scope}

  \draw[flowB] (3.02,5.34) -- (3.26,5.34);   
  \draw[flowG] (3.02,3.98) -- (3.26,3.98);   
  \draw[flowB] (11.06,6.60) -- (11.31,6.60); 
  \draw[flowG] (11.06,3.38) -- (11.20,3.38) -- (11.20,6.60);
\end{tikzpicture}
  \caption{The CAG framework. (a) Each appliance is routed once, from the duty cycle of its
  training windows: four UVIC appliances are always on and take the continuous branch, the
  other seven are event driven. (b) Intermittent appliances are partitioned into operating
  states by $k$-means over shape descriptors, with the idle cluster held out so that its
  mixture weight carries the duty cycle; one conditional generator and one multi-resolution
  discriminator are trained per cluster. The cluster centroids shown are the ones the
  clustering code finds for CoffeeMaker. (c) Continuous appliances are compressed to a load
  envelope by block averaging, modelled with an LSTM GAN, and reconstructed by repetition.
  (d) Cluster samples are drawn in proportion to the measured cluster shares and concatenated
  into a synthetic profile. (e) On top of this backbone, each appliance selects the adaptive
  components that address its own failure mode.}
  \label{fig:model_architecture}
\end{figure}

\subsection{Data Representation and Behavioral Routing}
Let $x_{1:T}\in\mathbb{R}^{T}$ denote the active-power trace of an appliance. For local load-pattern analysis, the trace is partitioned into non-overlapping fixed-length segments,
\begin{equation}
\mathcal{S}=\{s^{(j)}\in\mathbb{R}^{L}\}_{j=1}^{N},
\label{eq:segment-set}
\end{equation}
where $L$ is the segment length. The same raw trace is also examined globally to determine whether it should be treated as intermittent or continuous. This routing step is intentionally lightweight. Its purpose is not to build a detailed supervised appliance classifier, but to separate load profiles whose operating characteristics call for different synthesis strategies.

We compute three statistics from $x_{1:T}$. Let $\Delta x_t=x_t-x_{t-1}$ and let $\mathbb{1}[\cdot]$ denote the indicator function:
\begin{equation}
\begin{aligned}
R_0 &:= \mathbb{1}\big[\, x_1=\cdots=x_{\min\{T_0,\,T\}}=0\,\big],\\
p_{\mathrm{nz}} &:= \frac{1}{T}\sum_{t=1}^{T}\mathbb{1}[x_t\neq 0],\\
\sigma^2_{\Delta} &:= \frac{1}{T-1}\sum_{t=2}^{T}\big(\tilde{\Delta}x_t-\overline{\tilde{\Delta}x}\big)^2.
\end{aligned}
\label{eq:routing-stats}
\end{equation}
Here, $R_0$ captures traces that begin with an off-state prefix, $p_{\mathrm{nz}}$ measures the fraction of samples in which the appliance draws nonzero power, and $\sigma^2_{\Delta}$ measures short-term changes in active power after smoothing. The derivative $\tilde{\Delta}x_t$ is obtained by applying a 7-point moving average filter before finite differencing, which reduces sensitivity to isolated measurement noise while preserving genuine switching events. With prefix length $T_0$, occupancy threshold $\rho$, and derivative-variance threshold $\tau$, the routing rule is
\begin{equation}
\mathrm{route}(x_{1:T})=
\begin{cases}
\text{continuous}, & R_0=1\; \text{or}\; \big(p_{\mathrm{nz}}>\rho\ \wedge\ \sigma^2_{\Delta}<\tau\big),\\
\text{intermittent}, & \text{otherwise}.
\end{cases}
\label{eq:routing-rule}
\end{equation}
In the experiments, we use $T_0=100$, $\rho=0.7$, and $\tau=0.1$. The first condition accommodates traces that begin with an extended off-state interval before transitioning to sustained operation. The second condition characterizes a continuous load profile, in which the appliance draws power for most of the observation period and exhibits limited short-term variation in its smoothed active-power signal. Appliances that do not satisfy either criterion are classified as intermittent loads. Such devices are characterized by discrete load events, switching transients, and abrupt state transitions that are more effectively modeled as localized consumption patterns than as a single continuous load profile.

\subsection{Cluster Aggregation for Intermittent Patterns}
Intermittent appliances are challenging to synthesize because their load profiles jointly encode event timing, power magnitude, operating duration, and waveform characteristics. A coffee maker, microwave oven, or printer may exhibit one dominant operating cycle together with several less frequent cycles associated with warm-up, standby, user interruption, or partial operation. When all segments are used to train a single generator, the dominant operating cycle tends to dominate the learned distribution, while infrequent operating states are underrepresented in the generated data. CAG therefore partitions intermittent load segments into groups with similar electrical characteristics and assigns a dedicated generator to each group.

For each segment $s\in\mathbb{R}^{L}$, we compute a normalized segment
\begin{equation}
\tilde{s}=\frac{s-\mu}{\sigma},
\label{eq:seg-normalize}
\end{equation}
where $\mu$ and $\sigma$ are the segment mean and standard deviation. If $\sigma=0$, only centering is applied. The descriptor vector $\phi:\mathbb{R}^{L}\to\mathbb{R}^{d}$ summarizes the electrical shape of the segment from complementary views: power level, variability, temporal trend, spectral content, local morphology, and coarse waveform shape. Specifically, $\phi(s)$ concatenates
\begin{itemize}
  \item statistical descriptors: $\mu$, $\sigma$, $\mathrm{skew}(\tilde{s})=\tfrac{1}{L}\sum\tilde{s}^{3}$, and excess kurtosis $\mathrm{kurt}(\tilde{s})-3$;
  \item trend descriptor: slope $\beta=\mathrm{cov}(t,\tilde{s})/\mathrm{var}(t)$ for $t=1{:}L$;
  \item frequency descriptor: dominant frequency $f^*=\arg\max_f |[\mathcal{F}(\tilde{s})](f)|$ from the DFT magnitude;
  \item morphology descriptors: peak and valley counts $\sum\mathbb{1}[\tilde{s}>0.5]$ and $\sum\mathbb{1}[\tilde{s}<-0.5]$;
  \item roughness and energy descriptors: $\mathrm{var}(\Delta\tilde{s})$ and $\|\tilde{s}\|_2^2$;
  \item coarse shape samples: $\{\tilde{s}(\lfloor\alpha^{(k)}(L-1)\rfloor)\}_{k=1}^{20}$ for $\alpha^{(k)}=(k-1)/19$.
\end{itemize}
These descriptors intentionally rely on interpretable signal characteristics rather than learned latent representations during the clustering stage. The objective is to group load segments according to quantities that can be directly examined, such as average power consumption, variability, switching activity, dominant periodicity, and overall waveform structure. Consequently, each cluster can be interpreted as a distinct appliance operating state or load-event category rather than merely a numerical partition of the feature space.

\paragraph{K-means Clustering.}
After z-score standardization across all segments of a device, the feature vectors are clustered by k-means:
\begin{equation}
\min_{\{c^{(j)}\},\{\mu^{(k)}\}}\sum_{j=1}^{N}\left\|\phi(s^{(j)})-\mu^{(c^{(j)})}\right\|^{2},
\qquad c^{(j)}\in\{1,\dots,K\}.
\label{eq:kmeans}
\end{equation}
The number of clusters is selected by maximizing the mean silhouette score under an upper bound $K\leq\kappa$:
\begin{equation}
\bar{s}(K)=\frac{1}{N}\sum_{j=1}^{N}\frac{b(j)-a(j)}{\max\{a(j),b(j)\}},
\label{eq:silhouette}
\end{equation}
where $a(j)$ is the mean intra-cluster distance for segment $j$ and $b(j)$ is its minimum mean distance to another cluster. When the silhouette score is ill-defined, such as when too few valid segments are available, CAG falls back to a small admissible $K$. For cluster $k$, the resulting training set is
\begin{equation}
\mathcal{D}^{(k)}=\{s^{(j)}:c^{(j)}=k\},
\qquad
\pi_k=\frac{|\mathcal{D}^{(k)}|}{\sum_{\ell=1}^{K}|\mathcal{D}^{(\ell)}|}.
\label{eq:cluster-dataset}
\end{equation}
The empirical mixture weight $\pi_k$ is subsequently used during sampling, ensuring that the synthesized dataset preserves not only the electrical characteristics of the measured load-event categories but also their relative occurrence frequencies within the original appliance data.

\paragraph{Cluster-specific CNN GAN training.}
Each intermittent-load cluster is modeled using a compact one-dimensional convolutional GAN. Employing a dedicated GAN for each cluster provides a clear separation of synthesis responsibilities: one generator may reproduce short-duration load events, another sustained ON-state intervals, and another multi-stage operating cycles. This cluster-specific modeling strategy is the primary mechanism through which CAG preserves infrequent operating states. Rather than forcing a single generator to learn a heterogeneous mixture of appliance load signatures, each generator focuses on the load-event category represented by its assigned cluster.

For cluster $k$, let $G_k(\cdot;\theta_k):\mathbb{R}^{d_z}\to[-1,1]^L$ map a random latent vector $z\sim\mathcal{N}(0,I)$ to a synthetic load segment, and let $D_k(\cdot;\psi_k):[-1,1]^L\to(0,1)$ denote the corresponding discriminator that distinguishes measured and synthetic segments. Segments in $\mathcal{D}^{(k)}$ are min--max normalized to $[-1,1]$ before training and transformed back to their original power scale after generation. Both networks employ lightweight Conv1D architectures, which are well suited for short-duration load events because local convolutional kernels effectively capture switching transients, power ramps, and characteristic load signatures while remaining robust to minor variations in event timing.

The discriminator and generator losses for cluster $k$ are
\begin{align}
\mathcal{L}^{(D)}_k
&=\frac{1}{2}\mathbb{E}_{x\sim p^{\mathrm{real}}_k}\big[-\log D_k(x;\psi_k)\big]
+\frac{1}{2}\mathbb{E}_{z\sim\mathcal{N}}\big[-\log\big(1-D_k(G_k(z;\theta_k);\psi_k)\big)\big],
\label{eq:gan-D}\\
\mathcal{L}^{(G)}_k
&=\mathbb{E}_{z\sim\mathcal{N}}\big[-\log D_k(G_k(z;\theta_k);\psi_k)\big].
\label{eq:gan-G}
\end{align}
All cluster models use the same loss family and training settings, but their parameters are fitted to different operating-pattern categories.

\paragraph{Cluster-weighted intermittent synthesis.}
During synthesis, CAG first samples a cluster index $k\sim$\textsc{Categorical$( \pi_1,\dots,\pi_K)$} according to the observed pattern frequency, then samples $z$, generates $G_k(z)$, applies the inverse scaling for that cluster, and concatenates generated segments until the target trace length is reached.

\subsection{Per-Device Adaptive Components}
The routing-and-clustering backbone described above provides a strong baseline, but a fixed generator family cannot be optimal for every appliance. Different devices exhibit different failure modes: spiky on/off loads under-represent switching transients, steady-state loads suffer from marginal power drift, and long-memory loads lose slow envelope dynamics. CAG therefore augments the backbone with four adaptive components, each enabled per device by its statistical signature.

\paragraph{Stabilised cluster training.}
Each per-cluster GAN is trained with three stabilisation mechanisms. First, an exponential moving average (EMA) of the generator weights with decay $\beta_{\mathrm{ema}}=0.999$ is maintained during training and used for synthesis, which reduces high-frequency weight oscillation and yields smoother generated waveforms. Second, the discriminator operates at multiple temporal resolutions: it pools its intermediate activations at several scales and scores each pooled representation, so that both short transients and long-range structure are judged. Third, small clusters (fewer than a minimum effective size) are augmented by random time-shifts of their segments before training, which enlarges the effective sample size without changing the segment statistics. Optionally, a dilated convolutional stack with dilation rates such as $(1,2,4,8)$ is appended to the per-cluster generator to capture long-range correlations on devices whose clusters contain sustained steady-state intervals.

\paragraph{Behavior factorisation for spiky loads.}
For appliances with a strongly bimodal on/off signature (e.g., printers and desktops), the per-cluster generator is replaced by a behaviour-factorised variant. The generator produces two heads from its shared convolutional features: an activity head $a(\cdot)$ that outputs a per-sample on/off decision through a straight-through Gumbel-sigmoid,
\begin{equation}
a=\mathbb{1}\big[\sigma((\ell+\epsilon)/\tau)>0.5\big],
\qquad \epsilon\sim\mathrm{Logistic}(0,1),
\label{eq:gumbel-switch}
\end{equation}
and a magnitude head $m(\cdot)$ that outputs the power level while active. The synthesized segment is
\begin{equation}
x = a\odot\big(2m-1\big)+\big(1-a\big)\cdot(-1),
\label{eq:behavior-compose}
\end{equation}
in model space, where $a$ is a hard binary switch with straight-through gradients. This decomposition separates the timing of power draw from its magnitude, which improves event statistics (activation count, duty cycle, run length) for spiky devices.

\paragraph{Marginal moment matching.}
To calibrate the generated power distribution, a moment-matching term is added to the generator objective. With $x_r$ and $x_g$ denoting real and generated segments from the same cluster,
\begin{equation}
\mathcal{L}_{\mathrm{mom}}=\Big|\bar{x}_g-\bar{x}_r\Big|+\Big|\hat\sigma_g-\hat\sigma_r\Big|,
\label{eq:moment}
\end{equation}
where $\bar{x}$ and $\hat\sigma$ are the pooled mean and standard deviation. The term is weighted by a small coefficient $\lambda_{\mathrm{mom}}$ and directly targets the power-marginal Wasserstein metric, correcting residual bias in average consumption and fluctuation intensity that adversarial loss alone leaves unfixed.

\paragraph{Generator blending.}
When two generator families are complementary for the same appliance (e.g., a per-cluster generator strong on waveform fidelity and a waveform-transposed generator strong on spectral realism), CAG fuses them through blending. Two independent generators $G^{(a)}$ and $G^{(b)}$ are trained on the same clusters, and a synthesized segment is
\begin{equation}
x = \alpha\,G^{(a)}(z) + (1-\alpha)\,G^{(b)}(z'),
\label{eq:blend}
\end{equation}
with mixing weight $\alpha\in[0,1]$ chosen per device. Blending acts as an architectural ensemble over the two generator families. It is selected for Microwave, Laptop and iMac, where it lowers FID and event distance at the same time (Table~\ref{tab:ablation-study}).

\subsection{Continuous Sequence Generation}
Continuous appliances present a different load-modeling challenge. Their load profiles are typically characterized by sustained operation, gradual power variations, duty-cycle behavior, and relatively stable consumption levels. Consequently, the relevant information lies in the long-term evolution of the load profile rather than in discrete load events or switching transients. Training directly on full-resolution measurements introduces a large number of highly correlated steady-state samples, which can increase computational cost and complicate adversarial training. CAG therefore adopts a temporal aggregation and reconstruction strategy: the measured load profile is compressed before training, the resulting load envelope is modeled by a recurrent GAN, and the generated sequence is subsequently reconstructed to the original sampling resolution.

Given an averaging factor $F\geq1$, the raw sequence is partitioned into blocks of length $F$ and converted into a compressed load curve $\hat{x}$:
\begin{equation}
\hat{x}_i=\frac{1}{F}\sum_{t=(i-1)F+1}^{iF}x_t,
\quad i=1{:}\left\lfloor\frac{T}{F}\right\rfloor.
\label{eq:avg-simplify}
\end{equation}
This temporal aggregation suppresses redundant short-term fluctuations while preserving the dominant consumption trend and duty-cycle envelope of the active-power profile. When $|\hat{x}|$ exceeds a predefined budget $U$, overlapping windows are employed to ensure that the recurrent model processes sequences of manageable length.

The continuous branch employs an LSTM-based generator and discriminator. The generator transforms a latent vector into a time-varying load envelope through stacked LSTM layers followed by a linear $\tanh$ output layer. The discriminator receives either a measured or generated compressed load envelope and estimates its likelihood of originating from the measured data distribution. The adversarial objectives follow the same formulation as Eq.~\ref{eq:gan-D} and Eq.~\ref{eq:gan-G}, with $G_k,D_k$ replaced by the continuous generator and discriminator.

After generation, the compressed output $\hat{y}$ is returned to the original sampling horizon by block-wise replication:
\begin{equation}
\mathrm{Recon}(\hat{y},F)=
\big(\underbrace{\hat{y}_1,\dots,\hat{y}_1}_{F},\dots,
\underbrace{\hat{y}_n,\dots,\hat{y}_n}_{F}\big)\big|_{1:T},
\label{eq:reconstruction}
\end{equation}
where cropping or padding is applied when necessary. This reconstruction strategy is intuitively simple; the LSTM branch focuses on modeling long-term load dynamics and duty-cycle behavior, while the aggregation and replication operations maintain a stable and computationally efficient training process.

Algorithm~\ref{workflow} summarizes the complete training procedure. The important distinction from a standard GAN baseline is that CAG does not treat segmentation, routing, and clustering as incidental preprocessing. These steps define the load units that are synthesized, making the final synthetic dataset a composition of interpretable appliance operating patterns.

\begin{algorithm}
\caption{Adaptive CAG training pipeline}
\begin{algorithmic}[1]
\State Load the UVIC active-power traces and group them by appliance
\For{each appliance load profile $x_{1:T}$}
\State Compute routing statistics $R_0$, $p_{\mathrm{nz}}$, and $\sigma^2_{\Delta}$
\State Classify the profile as continuous or intermittent using Eq.~\ref{eq:routing-rule}
\If{intermittent}
\State Partition the profile into fixed-length load segments ${s^{(j)}}*{j=1}^{N}$
\State Extract electrical-feature descriptors $\phi(s^{(j)})$ and determine $K$ using the silhouette criterion
\State Cluster segments into ${\mathcal{D}^{(k)}}*{k=1}^{K}$ and estimate mixture weights ${\pi_k}_{k=1}^{K}$
\For{each cluster $k$}
\State Train a compact Conv1D GAN on $\mathcal{D}^{(k)}$
\EndFor
\Else
\State Compress the load profile using block averaging and optional windowing
\State Train an LSTM-GAN on the compressed load-envelope representation
\State Reconstruct generated load profiles to the original sampling horizon
\EndIf
\State Save trained generators, loss histories, synthesized samples, and clustering diagnostics
\EndFor
\end{algorithmic}
\label{workflow}
\end{algorithm}

\section{Experiments}\label{sec:exp}
\subsection{Dataset}
\begin{figure}
  \centering
  \includegraphics[width=0.95\linewidth]{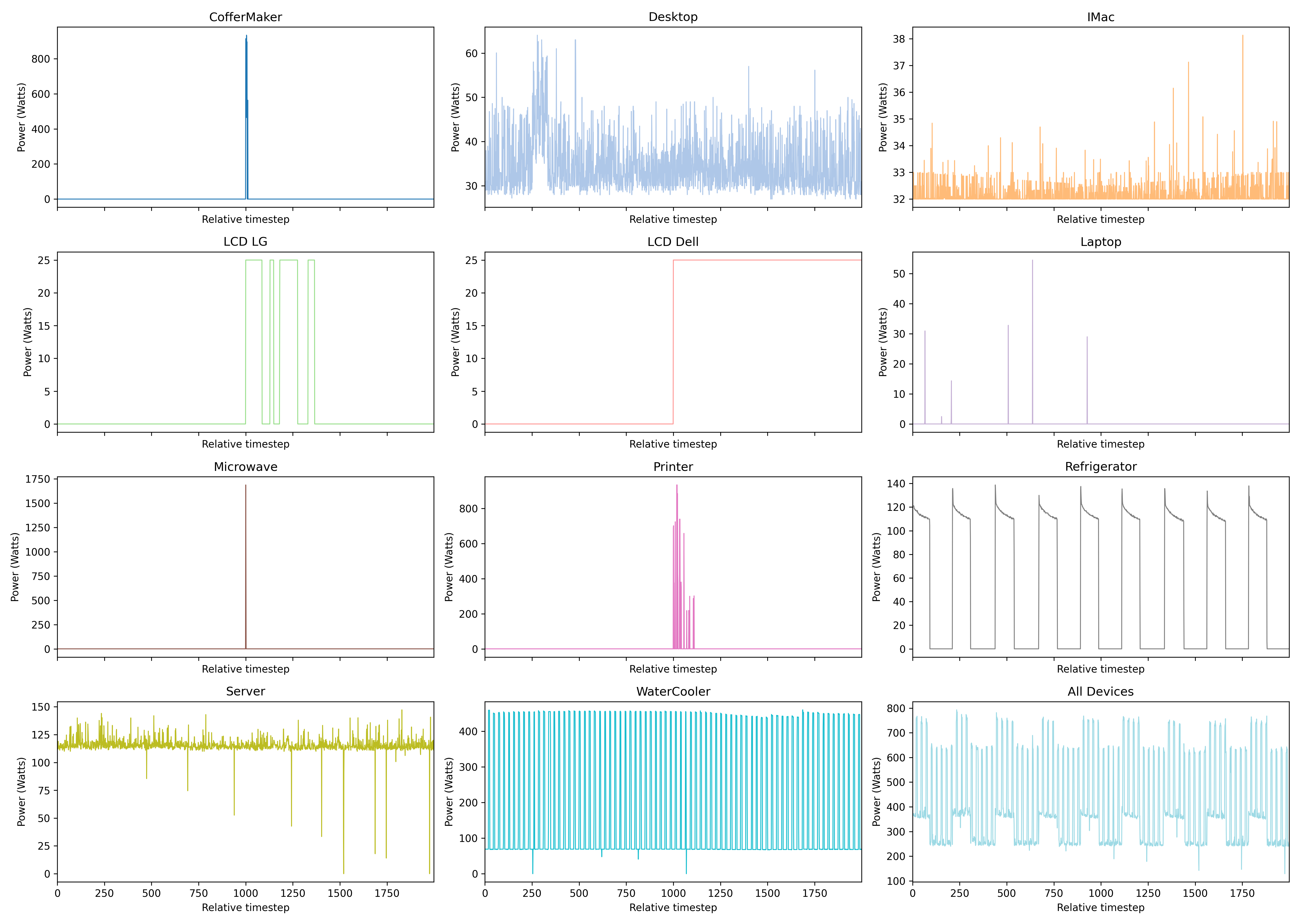}
  \caption{Visualization of appliance load profiles in the dataset.
Representative time series power consumption patterns for multiple household and office appliances from the UVIC dataset. Each subplot shows the active power over time for a specific device, illustrating the diversity of load behaviors across appliance types. Intermittent devices such as the CoffeeMaker, Microwave, and Printer exhibit short, sporadic activations, whereas continuous devices such as the Desktop, IMac, Server, and Refrigerator display long-duration, smoothly varying consumption trajectories. This variability in temporal structure and amplitude is the heterogeneity that the proposed CAG framework is built to model.}
\label{dataset_visualization}
\end{figure}

The experiments are conducted using the UVIC dataset, which provides appliance-level active-power measurements collected from residential and office environments. The dataset contains a diverse range of household and electronic appliances exhibiting distinct electricity-consumption characteristics. Figure~\ref{dataset_visualization} shows representative appliance load profiles. The recorded active-power signals capture the temporal behavior of individual appliances at high sampling resolution. Continuously operating appliances, such as desktop computers, iMacs, servers, and refrigerators, typically exhibit sustained consumption levels and gradual power variations, whereas intermittent appliances, including coffee makers, microwave ovens, printers, and water coolers, are characterized by discrete load events and frequent operating-state transitions.

The raw measurements were preprocessed to improve data quality and ensure consistency across appliances. The UVIC dataset comprises eleven appliance categories spanning a broad range of operating characteristics and electricity-consumption patterns. The presence of both continuous and intermittent loads makes the dataset suitable for evaluating appliance-level load-generation methods under diverse operating conditions. The processed data are subsequently used for training and validating the proposed CAG framework.

\subsection{Experimental settings}

The adaptive training pipeline of CAG is designed to assign appliance load profiles to appropriate generative branches and to facilitate adversarial learning across heterogeneous appliance categories. Each appliance is first classified as intermittent or continuous using the routing criteria defined in Eqs.~\ref{eq:routing-stats} and \ref{eq:routing-rule}. Specifically, a profile is classified as continuous when the initial portion of the signal contains a prolonged off-state interval ($R_0=1$), or when the nonzero occupancy exceeds $\rho=0.7$ and the variance of the 7-point smoothed derivative remains below $\tau=0.1$. Otherwise, the profile is classified as intermittent. This criterion is motivated by the observation that continuously operating appliances typically exhibit sustained power consumption with limited short-term variation, whereas intermittent appliances are characterized by discrete load events, switching transients, and frequent operating-state changes.

For intermittent appliances, load profiles are partitioned into fixed-length segments, standardized, and clustered using the electrical-feature descriptors described in Eq~\ref{eq:kmeans}. Each cluster corresponds to a distinct operating-state category and is modeled using an independent lightweight convolutional GAN. The generator and discriminator employ compact one-dimensional convolutional architectures with small receptive fields, enabling the learning of localized load signatures while maintaining computational efficiency. These convolutional operators are well suited for modeling short-duration load events because they capture switching transients, power ramps, and characteristic waveform structures while remaining robust to minor temporal misalignment.

For continuously operating appliances, long-duration load profiles are temporally aggregated and, when necessary, partitioned into overlapping windows before training. The compressed load-envelope representation is then modeled using a two-layer LSTM-based GAN. The recurrent architecture captures long-term temporal dependencies and generates representative load envelopes, which are subsequently reconstructed to the original sampling horizon using Eq.~\ref{eq:reconstruction}. This aggregation–reconstruction strategy reduces temporal dimensionality, improves computational efficiency, and facilitates stable adversarial training on long sequences.

The hyperparameter settings of the core pipeline are summarized in Table~\ref{tab:training-config}. Beyond these core settings, each appliance is assigned its best generative configuration by the adaptive components of Section~\ref{sec:method}: per-cluster generators with $K$ selected by silhouette under an upper bound (typically $K\in\{2,\dots,5\}$), optionally behaviour-factorised ($\tau=0.5$) with moment matching ($\lambda_{\mathrm{mom}}=50$), optionally dilated, optionally blended with a waveform-transposed generator ($\alpha$ per device). The per-device configuration is selected on the training fold and evaluated on the held-out fold, and all reported results are two-fold averages.

\begin{table}[t]
  \centering
  \caption{Core hyperparameters of the Cluster Aggregated GAN pipeline.}
  \label{tab:training-config}
  \renewcommand{\arraystretch}{0.95}
  \setlength{\tabcolsep}{4pt}
  \begin{tabularx}{\linewidth}{P{0.30\linewidth} Y}
    \toprule
    \multicolumn{2}{c}{\textbf{Common}}\\
    \midrule
    Segment length $L$        & 512 (85.3 minutes at 10\,s resolution) \\
    Latent dimension $d_z$    & 100 \\
    Training steps            & 6{,}000 per cluster \\
    Batch size                & 64 \\
    Optimiser                 & $\mathrm{Adam}(\eta=2\times10^{-4},\,\beta_1=0.5,\,\beta_2=0.9)$ \\
    EMA decay                 & 0.999 \\
    \midrule
    \multicolumn{2}{c}{\textbf{Per-cluster generator (intermittent branch)}}\\
    \midrule
    Generator channels        & 64 (ConvTranspose1d) \\
    Discriminator channels    & 48, multi-resolution scoring \\
    Augmentation              & time-shift for small clusters \\
    Optional dilated stack    & rates $(1,2,4,8)$ or $(2,4,8)$ \\
    Behavior factorisation    & Gumbel-sigmoid $\tau=0.5$, moment weight 50 \\
    \midrule
    \multicolumn{2}{c}{\textbf{Waveform-transposed branch}}\\
    \midrule
    Generator channels        & 64 (WaveGAN-style, PhaseShuffle) \\
    Critic updates            & 5 per generator step (WGAN-GP, $\lambda_{\mathrm{gp}}=10$) \\
    Learning rate             & $1\times10^{-4}$ \\
    \midrule
    \multicolumn{2}{c}{\textbf{Continuous branch}}\\
    \midrule
    Envelope factor $F$       & $\geq1$ (block averaging), horizon $\leq$ 1000 \\
    Generator / Discriminator & recurrent (LSTM), $\tanh$ output \\
    Reconstruction            & block replication, Eq.~\eqref{eq:reconstruction} \\
    \bottomrule
  \end{tabularx}
\end{table}

\subsection{Evaluation Metrics}
We benchmark all models with a set of domain-driven metrics that jointly capture the realism and diversity of appliance load curves. These two dimensions mirror the goals of practical NILM and data synthesis: realism requires that generated traces align with real power levels, waveform shapes, and duty cycles, whereas diversity requires that the generator covers the full variety of user behaviours. Let $\mathcal{X}_r=\{x^{(r)}\}$ and $\mathcal{X}_g=\{x^{(g)}\}$ denote real and generated sequences, and let $\Phi(\cdot)$ denote a feature extractor. We report eight metrics, grouped into realism, event, and diversity categories. Table~\ref{tab:gan-quality} summarises the average metrics over all devices, and Table~\ref{tab:device-model-quality} reports the detailed measurements for each device. Lower values are better for realism and event metrics; higher values are better for diversity metrics.

\subsubsection{Realism Metrics}
\begin{itemize}
  \item \textbf{Maximum Mean Discrepancy (MMD).} We measure the squared maximum mean discrepancy with an RBF kernel over a ladder of bandwidths derived from the real data,
\begin{equation*}
  \mathrm{MMD}^2=\frac{1}{B}\sum_{b=1}^{B}\Big[k_{xx}-2k_{xy}+k_{yy}\Big],
\end{equation*}
where $k_{xy}$ denotes the mean Gaussian kernel value with bandwidth $\sigma_b$ evaluated on feature embeddings of the real and generated sets. Because the bandwidths are fixed from the real sample only, all models are scored with the identical kernel, and lower values indicate closer waveform alignment.

  \item \textbf{Power-Marginal Wasserstein ($W_1$).} The 1-Wasserstein distance between the pooled marginal power distributions,
\begin{equation*}
  W_1=\inf_{\gamma\in\Pi(\hat p_r,\hat p_g)}\mathbb{E}_{(a,b)\sim\gamma}\left|a-b\right|,
\end{equation*}
where $\hat p_r$ and $\hat p_g$ are the empirical distributions of all power samples in the real and generated sets. This metric is the clearest indicator of whether the generator has recovered the long-term energy budget of a device; large deviations imply that downstream NILM models would consistently over- or under-estimate consumption even if the temporal structure looked plausible.

  \item \textbf{Autocorrelation Error (ACF).} Temporal consistency is measured through the mean absolute difference of the autocorrelation functions up to lag $\tau_{\max}$,
\begin{equation*}
  \mathrm{ACF}=\frac{1}{\tau_{\max}}\sum_{\tau=1}^{\tau_{\max}}\Big|\hat\rho_r(\tau)-\hat\rho_g(\tau)\Big|,
\qquad
\hat\rho(\tau)=\frac{\frac{1}{L-\tau}\sum_{t=1}^{L-\tau}(x_t-\bar x)(x_{t+\tau}-\bar x)}{\mathrm{Var}(x)}.
\end{equation*}
Low values imply that generated traces preserve the persistence structure of the real signal, which matters for appliances whose duty cycles and ramp shapes are defined by correlated successive samples.

  \item \textbf{Power Spectral Density Error (PSD).} Frequency-domain realism is evaluated on the mean absolute log-spectral difference,
\begin{equation*}
  \mathrm{PSD}=\frac{1}{N_f}\sum_{k}\Big|\log_{10}\hat p_r(f_k)-\log_{10}\hat p_g(f_k)\Big|,
\end{equation*}
where $\hat p(f)$ is the mean periodogram over windows. This metric penalises models that match marginal statistics but place energy at the wrong frequencies (e.g., overly smooth or excessively oscillatory traces).

  \item \textbf{Fréchet Inception-style Distance (FID).} Structural similarity over learned waveform descriptors is measured through the Fréchet distance in the feature space of an appliance classifier $\Phi$,
\begin{equation*}
  \mathrm{FID}=\lVert m_r-m_g\rVert_2^2+\mathrm{Tr}\big(C_r+C_g-2(C_r^{1/2}C_gC_r^{1/2})^{1/2}\big),
\end{equation*}
with $(m_r,C_r)$ and $(m_g,C_g)$ the empirical means and covariances of $\Phi(x^{(r)})$ and $\Phi(x^{(g)})$. Smaller values correspond to higher structural realism, capturing correlated variations such as multi-stage activation patterns and sustained operating levels.
\end{itemize}

\subsubsection{Event Metric}
\begin{itemize}
  \item \textbf{Event Statistics Distance (Evt).} For intermittent appliances, we extract four per-window event statistics from the binarised power signal $x>\theta$: duty cycle, activation count, mean ON power, and mean ON run length. The metric is the mean of four normalised 1-Wasserstein distances over these statistics,
\begin{equation*}
  \mathrm{Evt}=\frac{1}{4}\sum_{s\in\{\mathrm{duty},n_{\mathrm{ev}},p_{\mathrm{on}},r_{\mathrm{on}}\}}\frac{W_1(\hat p_r^{(s)},\hat p_g^{(s)})}{Z_s},
\end{equation*}
where $Z_s$ normalises each statistic by its natural scale. This replaces the earlier period-MAE heuristic, which matched each generated trace to its nearest real trace in dominant frequency and thereby biased the score towards zero by construction; the event-statistic distance instead compares full distributions of switching behaviour.
\end{itemize}

\subsubsection{Diversity Metrics}
\begin{itemize}
  \item \textbf{Density and Coverage.} Following Naeem et al.~\cite{naeem2020reliable}, we compute the $k$-NN density and coverage in the feature space $\Phi$,
\begin{equation*}
  \mathrm{Density}=\frac{1}{kN_g}\sum_{g}\mathbf{1}\big[x^{(g)}\in \mathcal{N}_k(x^{(r)})\big],\qquad
  \mathrm{Coverage}=\frac{1}{N_r}\sum_{r}\mathbf{1}\big[\mathcal{N}_k(x^{(r)})\cap \mathcal{X}_g\neq\emptyset\big],
\end{equation*}
where $\mathcal{N}_k(x)$ denotes the $k$-nearest-neighbour ball of a real sample. Density measures how many generated samples fall inside the neighbourhoods of real samples (fidelity), while Coverage measures the fraction of real modes reached by the generator (diversity). A mode-collapsed model scores high Density with near-zero Coverage, so the pair must be read together.

  \item \textbf{Joint Quality (F1).} To summarise the trade-off, we report the harmonic mean of Density and Coverage,
\begin{equation*}
  F_1=\frac{2\cdot\mathrm{Density}\cdot\mathrm{Coverage}}{\mathrm{Density}+\mathrm{Coverage}},
\end{equation*}
which is used as the ranking criterion for the fidelity--diversity pair, preventing collapsed baselines from winning on raw density alone. Raw Density and Coverage values are still reported in the tables.
\end{itemize}

All models are evaluated on the held-out test fold with 512 generated samples per device, and results are averaged over two train/test folds. Because a GAN that collapses to a constant output produces pathologically low $W_1$ and ACF scores while achieving zero coverage, we additionally exclude collapsed runs (Coverage $<0.005$ and Density $<0.005$) from the baseline statistics; this prevents training failures from being mistaken for strong realism.

\subsection{Baselines}
CAG is evaluated against seven GAN-based time-series generation models: CNN-GAN, LSTM-GAN, RNN-GAN, WaveGAN~\citep{yang2022wavegan}, BehaviorGAN~\citep{Castangia2025_BehaviorGAN}, TraceGAN~\citep{Harell2021TraceGAN}, and RLP-GAN~\citep{Liang2025GANPatternsTCE}. A routing-and-clustering-only variant of the proposed framework is the ablation reference. That variant models intermittent clusters with independent CNN GANs and continuous loads with an LSTM GAN, without the per-device adaptive components introduced in this work. All methods are trained on the UVIC dataset under identical preprocessing procedures, optimization settings, and sampling budgets. This comparison assesses the effectiveness of the proposed per-device adaptive generation strategy relative to conventional convolutional, recurrent, waveform-based, and behaviour-aware generative models.

\section{Results and Discussion}\label{sec:result}
\begin{figure}
  \centering
  \includegraphics[width=0.95\linewidth]{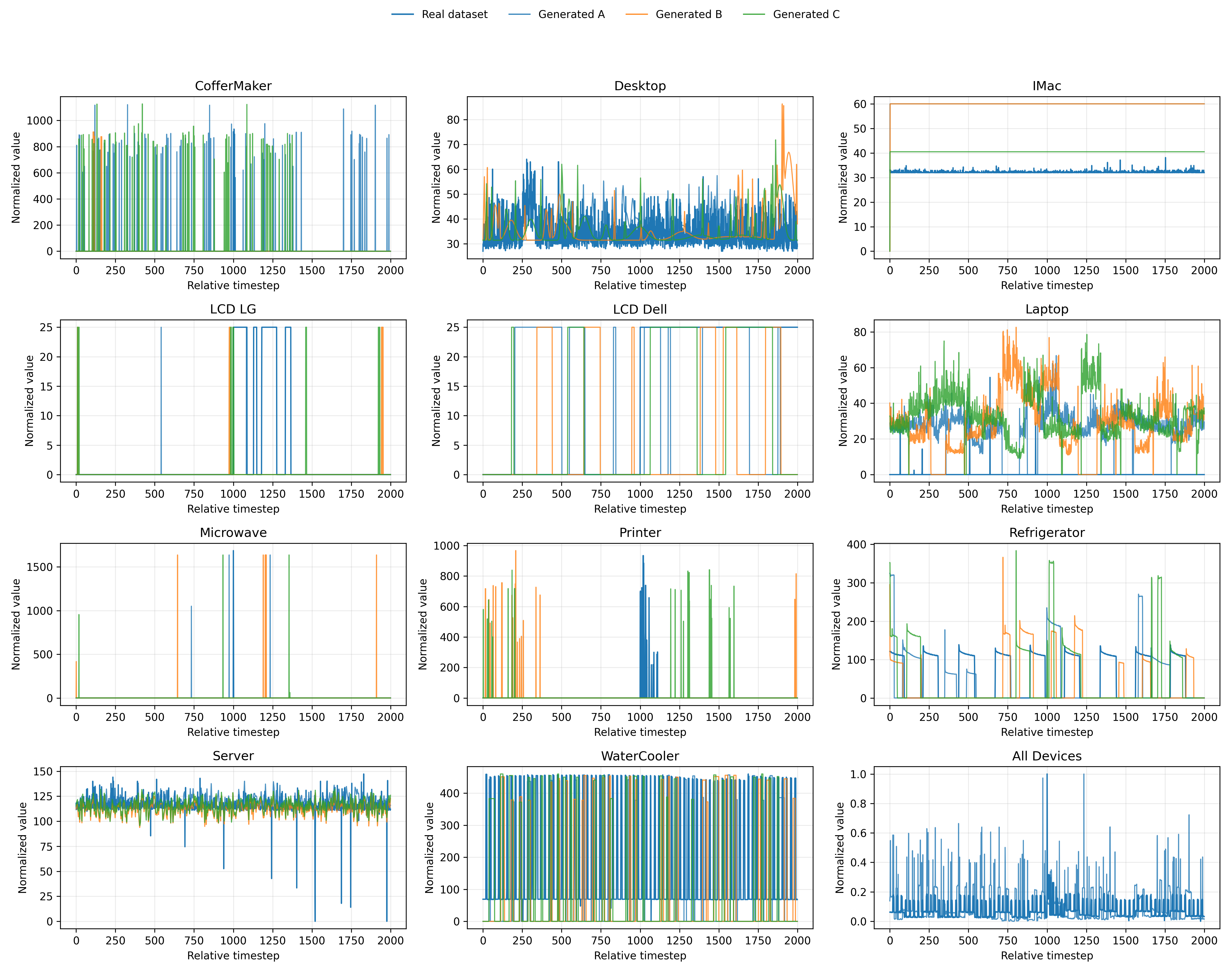}
  \caption{Example outputs of the proposed CAG framework. For each appliance, three generated traces (Generated A, B, and C) are overlaid on a measured reference trace from the UVIC dataset. The examples show whether the generated load curves preserve the visible operating features of each appliance, including burst timing for intermittent loads and smooth load envelopes for continuous devices.}
  \label{fig:cag_output}
\end{figure}

Using the baselines described above, we compare CAG against CNN-GAN, LSTM-GAN, RNN-GAN, WaveGAN, BehaviorGAN, TraceGAN, and RLP-GAN under identical training conditions on the UVIC dataset. Table~\ref{tab:gan-quality} reports average realism, event, and diversity metrics across all appliances, and Figure~\ref{fig:cag_output} presents representative generated load profiles. Table~\ref{tab:device-model-quality} provides the device-level results. There, \triup\ marks a ranking metric on which CAG matches or beats every baseline and \tridown\ one on which it does not, and the last column reports how many of the eight ranking metrics CAG wins per device. The primary objective is not only to reproduce the average energy consumption of an appliance but also to preserve the temporal and operational characteristics that define its behavior, including switching events, duty-cycle patterns, and operating-state transitions.

\begin{table}[t]
  \centering
  \renewcommand{\arraystretch}{0.9}
  \setlength{\tabcolsep}{2pt}
  \caption{Average realism, event, and diversity metrics across all evaluated devices . Bold marks the best non-oracle value; $F_1$ is the harmonic mean of Density and Coverage and is the ranking criterion for the fidelity--diversity pair.}
  \label{tab:gan-quality}
\begin{tabular}{cccccccccc}
  \toprule
  \multirow{2}{*}{\textbf{Model}}
    & \multicolumn{5}{c}{\textbf{Realism}}
    & \textbf{Event}
    & \multicolumn{3}{c}{\textbf{Diversity}} \\
  \cmidrule(lr){2-6} \cmidrule(lr){7-7} \cmidrule(lr){8-10}
  & \textbf{MMD $\downarrow$} & \textbf{$W_1$ $\downarrow$} & \textbf{ACF $\downarrow$} & \textbf{PSD $\downarrow$} & \textbf{FID $\downarrow$}
    & \textbf{Evt $\downarrow$}
    & \textbf{Den $\uparrow$} & \textbf{Cov $\uparrow$} & \textbf{$F_1$ $\uparrow$} \\
  \midrule
  CNN-GAN & $0.274$\tridown & $0.490$\tridown & $0.106$\tridown & $1.445$\tridown & $551.5$\tridown & $0.357$\tridown & $0.186$\tridown & $0.046$\tridown & $0.059$\tridown \\
  LSTM-GAN & $0.277$\tridown & $0.341$\tridown & $0.095$\tridown & $1.909$\tridown & $749.3$\tridown & $0.187$\tridown & $0.195$\tridown & $0.017$\tridown & $0.030$\tridown \\
  RNN-GAN & $0.460$\tridown & $1.150$\tridown & $0.085$\tridown & $2.922$\tridown & $794.7$\tridown & $0.258$\tridown & $0.161$\tridown & $0.018$\tridown & $0.026$\tridown \\
  WaveGAN & $0.130$\tridown & $0.182$\tridown & $0.055$\tridown & $2.650$\tridown & $363.4$\tridown & $0.109$\tridown & $\mathbf{4.627}$\triup & $0.254$\tridown & $0.365$\tridown \\
  BehaviorGAN & $0.235$\tridown & $0.370$\tridown & $0.088$\tridown & $2.543$\tridown & $652.1$\tridown & $0.110$\tridown & $3.705$\tridown & $0.169$\tridown & $0.287$\tridown \\
  TraceGAN & $0.278$\tridown & $1.121$\tridown & $0.084$\tridown & $1.308$\tridown & $1107.3$\tridown & $0.547$\tridown & $0.131$\tridown & $0.043$\tridown & $0.056$\tridown \\
  RLP-GAN & $0.435$\tridown & $0.450$\tridown & $0.097$\tridown & $2.062$\tridown & $883.8$\tridown & $0.221$\tridown & $0.145$\tridown & $0.010$\tridown & $0.018$\tridown \\
  CAG & $\mathbf{0.041}$\triup & $\mathbf{0.131}$\triup & $\mathbf{0.033}$\triup & $\mathbf{0.477}$\triup & $\mathbf{49.4}$\triup & $\mathbf{0.079}$\triup & $2.377$\tridown & $\mathbf{0.511}$\triup & $\mathbf{0.652}$\triup \\
  \bottomrule
\end{tabular}

\end{table}

The aggregate results in Table~\ref{tab:gan-quality} show that CAG holds the best average value on eight of the nine reported metrics. It reaches the lowest MMD (0.041), power-marginal Wasserstein distance (0.131), autocorrelation distance (0.033), spectral distance (0.477), FID (49.4) and event distance (0.079), together with the highest Coverage (0.511) and $F_1$ (0.652). Density is the exception, where WaveGAN reaches 4.627 against 2.377 for CAG. The Coverage value indicates that the per-device adaptive configuration reaches a broader range of appliance operating behaviors rather than repeatedly generating a small set of highly probable patterns. This balance between fidelity and diversity is particularly important for synthetic load generation because overly conservative generators often achieve low reconstruction error at the expense of behavioral coverage.

\begin{figure}
  \centering
  \includegraphics[width=\linewidth]{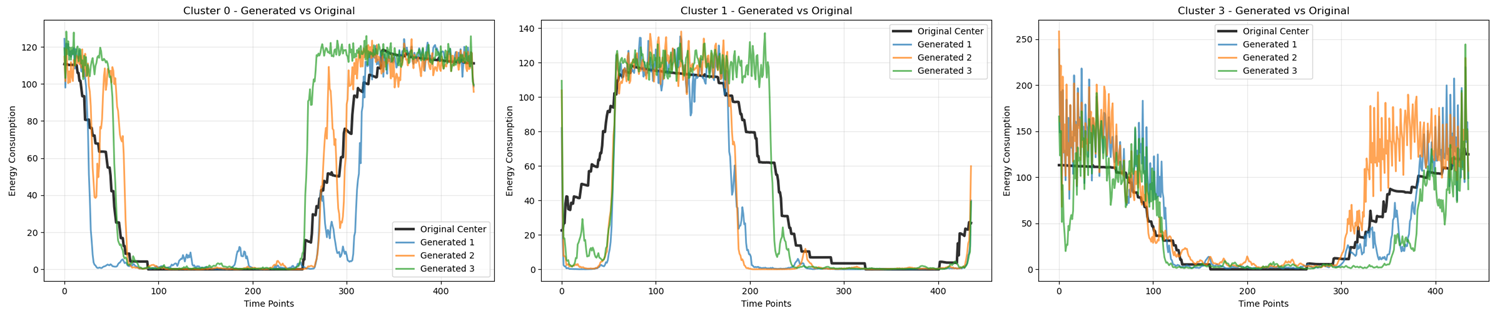}
  \caption{Per-cluster generation for an intermittent appliance. For three representative clusters, the black curve is the measured cluster centre and the coloured curves are independent samples drawn from the cluster-specific generator. The generated samples retain the characteristic activation shape of their own cluster, illustrating how CAG preserves distinct operating motifs rather than averaging them into one generic waveform.}
  \label{fig:cluster_generation}
\end{figure}

The observed performance gains can be linked directly to the design of the proposed framework. Conventional GAN architectures learn a single generative distribution for all operating conditions of an appliance. However, appliance load profiles are often multimodal. Intermittent devices such as coffee makers, printers, and microwave ovens exhibit multiple operating states, each characterized by distinct power levels, durations, and transition structures. Figure~\ref{fig:cluster_generation} illustrates that the clustering stage separates these operating states into more homogeneous load-event categories before adversarial learning. Consequently, each cluster-specific generator focuses on a narrower and more coherent distribution, reducing the tendency to average disparate operating patterns into a single generic waveform. The effect is clearest on the intermittent appliances: CAG wins eight of the eight ranking metrics on Coffee Maker and Water Cooler and seven on Printer (Table~\ref{tab:device-model-quality}).

A similar advantage is observed for continuously operating appliances. Devices such as desktops, servers, LCD displays, and refrigerators are characterized by long-duration load profiles containing extended steady-state intervals and gradual power variations. Direct adversarial training on full-resolution sequences introduces substantial temporal redundancy and increases optimization difficulty. The proposed aggregation–reconstruction strategy addresses this challenge by compressing the load profile into a lower-dimensional load-envelope representation prior to training. The LSTM-based branch therefore focuses on modeling long-term consumption dynamics rather than individual sample-to-sample fluctuations. CAG wins eight of the eight ranking metrics on Desktop, Server and Refrigerator and seven on iMac (Table~\ref{tab:device-model-quality}), which suggests that this compressed representation preserves the dominant operating characteristics.

The device-level results in Table~\ref{tab:device-model-quality} further support this interpretation. Although individual baselines occasionally outperform CAG on isolated metrics, these improvements rarely generalize across the full set of evaluation criteria. For example, some models achieve competitive waveform fidelity or variance matching for particular appliances, but these gains are often accompanied by substantial degradation in power-level accuracy, duty-cycle preservation, or diversity. In contrast, CAG consistently ranks near the top across multiple realism and diversity measures simultaneously. This consistency is particularly important for NILM and appliance-load modeling applications, where accurate representation of only one characteristic of the signal is insufficient. Synthetic data must preserve the joint relationship between power magnitude, temporal structure, operating-state transitions, and occurrence frequency to remain representative of real appliance behavior.

The visual examples shown in Figure~\ref{fig:cluster_generation} provide qualitative support for the quantitative findings. For intermittent appliances, the generated traces reproduce characteristic load events, activation durations, and transition structures observed in the measured signals. For continuous appliances, the generated profiles maintain the overall load envelope and long-term consumption trend without introducing unrealistic fluctuations. The qualitative agreement observed across the eleven appliance categories indicates that the proposed routing strategy successfully directs each device to a generative model aligned with its underlying operating behavior.

In summary, the results suggest that the advantage of CAG comes from incorporating appliance operating characteristics into the generation process rather than from added model capacity. By separating intermittent and continuous load behaviors and further decomposing intermittent loads into operating-state categories, CAG transforms a heterogeneous appliance-generation problem into a collection of simpler and more physically meaningful modeling tasks. The resulting synthetic load profiles exhibit improved realism, broader behavioral coverage, and stronger preservation of appliance operating structure, making them well suited for applications such as NILM development, synthetic data augmentation, privacy-preserving data sharing, and energy-consumption modeling.

{\renewcommand{\arraystretch}{0.9}\setlength{\tabcolsep}{2pt}\small
\begin{longtable}{cccccccccc}
  \caption{Per-device comparison of CAG against every baseline on the eight ranking metrics (two-fold average). \triup\ marks the best model for that device and metric, \tridown\ every other model; the count after each device name is the number of ranking metrics on which CAG is best.}
  \label{tab:device-model-quality} \\
  \toprule
  \multirow{2}{*}{\textbf{Device (SOTA)}} & \multirow{2}{*}{\textbf{Model}}
    & \multicolumn{5}{c}{\textbf{Realism}}
    & \textbf{Event}
    & \multicolumn{2}{c}{\textbf{Diversity}} \\
  \cmidrule(lr){3-7} \cmidrule(lr){8-8} \cmidrule(lr){9-10}
  & & \textbf{MMD $\downarrow$} & \textbf{$W_1$ $\downarrow$} & \textbf{ACF $\downarrow$} & \textbf{PSD $\downarrow$} & \textbf{FID $\downarrow$}
    & \textbf{Evt $\downarrow$}
    & \textbf{$F_1$ $\uparrow$} & \textbf{Cov $\uparrow$} \\
  \midrule
  \endfirsthead
  \multicolumn{10}{l}{\textit{Table \thetable\ (continued)}} \\
  \toprule
  \multirow{2}{*}{\textbf{Device (SOTA)}} & \multirow{2}{*}{\textbf{Model}}
    & \multicolumn{5}{c}{\textbf{Realism}}
    & \textbf{Event}
    & \multicolumn{2}{c}{\textbf{Diversity}} \\
  \cmidrule(lr){3-7} \cmidrule(lr){8-8} \cmidrule(lr){9-10}
  & & \textbf{MMD $\downarrow$} & \textbf{$W_1$ $\downarrow$} & \textbf{ACF $\downarrow$} & \textbf{PSD $\downarrow$} & \textbf{FID $\downarrow$}
    & \textbf{Evt $\downarrow$}
    & \textbf{$F_1$ $\uparrow$} & \textbf{Cov $\uparrow$} \\
  \midrule
  \endhead
  \bottomrule
  \endlastfoot
  \multirow{8}{*}{Coffee Maker (8/8)} & CNN-GAN & $0.256$\tridown & $0.095$\tridown & $0.027$\tridown & $2.719$\tridown & $2575.1$\tridown & $0.071$\tridown & $0.078$\tridown & $0.042$\tridown \\
   & LSTM-GAN & $0.258$\tridown & $0.110$\tridown & $0.027$\tridown & $1.627$\tridown & $2545.8$\tridown & $0.108$\tridown & $0.067$\tridown & $0.035$\tridown \\
   & RNN-GAN & $0.261$\tridown & $0.092$\tridown & $0.029$\tridown & $5.507$\tridown & $2583.3$\tridown & $0.066$\tridown & $0.067$\tridown & $0.035$\tridown \\
   & WaveGAN & $0.254$\tridown & $0.089$\tridown & $0.031$\tridown & $5.165$\tridown & $2550.2$\tridown & $0.077$\tridown & $0.089$\tridown & $0.048$\tridown \\
   & BehaviorGAN & $0.249$\tridown & $0.092$\tridown & $0.031$\tridown & $5.255$\tridown & $2271.7$\tridown & $0.060$\tridown & $0.095$\tridown & $0.052$\tridown \\
   & TraceGAN & $0.109$\tridown & $0.320$\tridown & $0.065$\tridown & $1.247$\tridown & $2368.1$\tridown & $0.188$\tridown & $0.120$\tridown & $0.072$\tridown \\
   & RLP-GAN & $0.282$\tridown & $0.079$\tridown & $0.032$\tridown & $2.692$\tridown & $2565.9$\tridown & $0.128$\tridown & $0.067$\tridown & $0.035$\tridown \\
   & CAG & $\mathbf{0.004}$\triup & $\mathbf{0.043}$\triup & $\mathbf{0.017}$\triup & $\mathbf{0.173}$\triup & $\mathbf{133.3}$\triup & $\mathbf{0.030}$\triup & $\mathbf{0.471}$\triup & $\mathbf{0.519}$\triup \\
  \midrule
  \multirow{8}{*}{Microwave (5/8)} & CNN-GAN & $0.243$\tridown & $0.038$\tridown & $0.034$\tridown & $3.724$\tridown & $208.3$\tridown & $0.037$\tridown & $0.013$\tridown & $0.007$\tridown \\
   & LSTM-GAN & $0.270$\tridown & $0.081$\tridown & $\mathbf{0.015}$\triup & $1.010$\tridown & $208.4$\tridown & $0.221$\tridown & $0.013$\tridown & $0.007$\tridown \\
   & RNN-GAN & $0.254$\tridown & $0.044$\tridown & $0.017$\tridown & $4.474$\tridown & $208.7$\tridown & $0.175$\tridown & $0.013$\tridown & $0.007$\tridown \\
   & WaveGAN & $0.248$\tridown & $0.032$\tridown & $0.020$\tridown & $8.080$\tridown & $208.2$\tridown & $\mathbf{0.034}$\triup & $0.013$\tridown & $0.007$\tridown \\
   & BehaviorGAN & $0.248$\tridown & $\mathbf{0.032}$\triup & $0.019$\tridown & $5.962$\tridown & $209.5$\tridown & $0.045$\tridown & $0.013$\tridown & $0.007$\tridown \\
   & TraceGAN & $0.132$\tridown & $0.260$\tridown & $0.041$\tridown & $1.076$\tridown & $204.0$\tridown & $1.269$\tridown & $0.017$\tridown & $0.038$\tridown \\
   & RLP-GAN & $0.281$\tridown & $0.058$\tridown & $0.019$\tridown & $2.965$\tridown & $208.2$\tridown & $0.225$\tridown & $0.013$\tridown & $0.007$\tridown \\
   & CAG & $\mathbf{0.006}$\triup & $0.035$\tridown & $0.016$\tridown & $\mathbf{0.336}$\triup & $\mathbf{24.4}$\triup & $0.053$\tridown & $\mathbf{0.196}$\triup & $\mathbf{0.191}$\triup \\
  \midrule
  \multirow{8}{*}{Laptop (6/8)} & CNN-GAN & $0.139$\tridown & $0.233$\tridown & $0.187$\tridown & $0.515$\tridown & $185.1$\tridown & $0.498$\tridown & $0.211$\tridown & $0.187$\tridown \\
   & LSTM-GAN & $0.099$\tridown & $0.180$\tridown & $0.207$\tridown & $1.097$\tridown & $578.6$\tridown & $0.178$\tridown & $0.057$\tridown & $0.031$\tridown \\
   & RNN-GAN & $0.067$\tridown & $0.143$\tridown & $0.131$\tridown & $1.120$\tridown & $530.8$\tridown & $0.632$\tridown & $0.051$\tridown & $0.047$\tridown \\
   & WaveGAN & $0.055$\tridown & $0.151$\tridown & $0.142$\tridown & $0.516$\tridown & $68.7$\tridown & $\mathbf{0.169}$\triup & $0.310$\tridown & $0.292$\tridown \\
   & BehaviorGAN & $0.237$\tridown & $0.476$\tridown & $0.192$\tridown & $0.847$\tridown & $336.0$\tridown & $0.230$\tridown & $0.066$\tridown & $0.037$\tridown \\
   & TraceGAN & $0.066$\tridown & $0.293$\tridown & $\mathbf{0.073}$\triup & $0.417$\tridown & $445.6$\tridown & $0.902$\tridown & $0.200$\tridown & $0.140$\tridown \\
   & RLP-GAN & $0.428$\tridown & $0.551$\tridown & $0.206$\tridown & $0.346$\tridown & $514.3$\tridown & $0.512$\tridown & $0.000$\tridown & $0.000$\tridown \\
   & CAG & $\mathbf{0.049}$\triup & $\mathbf{0.097}$\triup & $0.095$\tridown & $\mathbf{0.278}$\triup & $\mathbf{62.1}$\triup & $0.208$\tridown & $\mathbf{0.413}$\triup & $\mathbf{0.416}$\triup \\
  \midrule
  \multirow{8}{*}{Desktop (8/8)} & CNN-GAN & $0.059$\tridown & $0.374$\tridown & $0.068$\tridown & $0.431$\tridown & $179.2$\tridown & $0.010$\tridown & $0.051$\tridown & $0.030$\tridown \\
   & LSTM-GAN & $0.175$\tridown & $0.552$\tridown & $0.029$\tridown & $3.750$\tridown & $1992.7$\tridown & $0.006$\tridown & $0.000$\tridown & $0.000$\tridown \\
   & RNN-GAN & $0.591$\tridown & $2.361$\tridown & $0.043$\tridown & $3.112$\tridown & $830.3$\tridown & $0.437$\tridown & $0.009$\tridown & $0.016$\tridown \\
   & WaveGAN & $0.053$\tridown & $0.285$\tridown & $0.030$\tridown & $0.383$\tridown & $103.7$\tridown & $0.003$\tridown & $0.164$\tridown & $0.119$\tridown \\
   & BehaviorGAN & $0.460$\tridown & $1.056$\tridown & $0.136$\tridown & $0.842$\tridown & $2396.3$\tridown & $0.036$\tridown & $0.000$\tridown & $0.000$\tridown \\
   & TraceGAN & $0.364$\tridown & $1.161$\tridown & $0.068$\tridown & $1.095$\tridown & $2644.8$\tridown & $0.043$\tridown & $0.000$\tridown & $0.000$\tridown \\
   & RLP-GAN & $0.373$\tridown & $0.713$\tridown & $0.055$\tridown & $1.501$\tridown & $1625.0$\tridown & $0.006$\tridown & $0.000$\tridown & $0.000$\tridown \\
   & CAG & $\mathbf{0.004}$\triup & $\mathbf{0.069}$\triup & $\mathbf{0.011}$\triup & $\mathbf{0.098}$\triup & $\mathbf{7.4}$\triup & $\mathbf{0.002}$\triup & $\mathbf{0.585}$\triup & $\mathbf{0.590}$\triup \\
  \midrule
  \multirow{8}{*}{Printer (7/8)} & CNN-GAN & $0.157$\tridown & $0.163$\tridown & $0.052$\tridown & $1.733$\tridown & $673.8$\tridown & $0.134$\tridown & $0.051$\tridown & $0.029$\tridown \\
   & LSTM-GAN & $0.223$\tridown & $0.101$\tridown & $0.025$\tridown & $1.362$\tridown & $667.5$\tridown & $0.104$\tridown & $0.052$\tridown & $0.029$\tridown \\
   & RNN-GAN & $0.209$\tridown & $0.054$\tridown & $0.025$\tridown & $7.367$\tridown & $668.4$\tridown & $0.046$\tridown & $0.062$\tridown & $0.034$\tridown \\
   & WaveGAN & $0.209$\tridown & $0.054$\tridown & $0.025$\tridown & $8.137$\tridown & $668.4$\tridown & $0.046$\tridown & $0.062$\tridown & $0.034$\tridown \\
   & BehaviorGAN & $0.133$\tridown & $0.053$\tridown & $0.025$\tridown & $4.320$\tridown & $570.3$\tridown & $\mathbf{0.039}$\triup & $0.285$\tridown & $0.255$\tridown \\
   & TraceGAN & $0.099$\tridown & $0.255$\tridown & $0.051$\tridown & $1.213$\tridown & $593.9$\tridown & $0.197$\tridown & $0.201$\tridown & $0.168$\tridown \\
   & RLP-GAN & $0.313$\tridown & $0.062$\tridown & $0.026$\tridown & $0.770$\tridown & $660.8$\tridown & $0.185$\tridown & $0.062$\tridown & $0.034$\tridown \\
   & CAG & $\mathbf{0.009}$\triup & $\mathbf{0.043}$\triup & $\mathbf{0.021}$\triup & $\mathbf{0.250}$\triup & $\mathbf{112.1}$\triup & $0.041$\tridown & $\mathbf{0.511}$\triup & $\mathbf{0.449}$\triup \\
  \midrule
  \multirow{8}{*}{iMac (7/8)} & CNN-GAN & $0.265$\tridown & $0.597$\tridown & $0.074$\tridown & $0.619$\tridown & $67.0$\tridown & $0.013$\tridown & $0.010$\tridown & $0.008$\tridown \\
   & LSTM-GAN & $0.126$\tridown & $0.586$\tridown & $0.019$\tridown & $1.160$\tridown & $36.3$\tridown & $0.014$\tridown & $0.025$\tridown & $0.019$\tridown \\
   & RNN-GAN & $0.142$\tridown & $0.215$\tridown & $\mathbf{0.018}$\triup & $1.291$\tridown & $17.8$\tridown & $0.010$\tridown & $0.012$\tridown & $0.017$\tridown \\
   & WaveGAN & $0.256$\tridown & $0.178$\tridown & $0.069$\tridown & $0.950$\tridown & $8.9$\tridown & $0.009$\tridown & $0.016$\tridown & $0.025$\tridown \\
   & BehaviorGAN & $0.349$\tridown & $0.346$\tridown & $0.039$\tridown & $1.547$\tridown & $34.7$\tridown & $0.012$\tridown & $0.028$\tridown & $0.018$\tridown \\
   & TraceGAN & $0.662$\tridown & $3.973$\tridown & $0.073$\tridown & $0.822$\tridown & $1593.6$\tridown & $0.050$\tridown & $0.000$\tridown & $0.000$\tridown \\
   & RLP-GAN & $0.753$\tridown & $0.256$\tridown & $0.040$\tridown & $1.099$\tridown & $17.6$\tridown & $0.009$\tridown & $0.000$\tridown & $0.000$\tridown \\
   & CAG & $\mathbf{0.047}$\triup & $\mathbf{0.142}$\triup & $0.021$\tridown & $\mathbf{0.209}$\triup & $\mathbf{1.6}$\triup & $\mathbf{0.009}$\triup & $\mathbf{0.377}$\triup & $\mathbf{0.321}$\triup \\
  \midrule
  \multirow{8}{*}{Server (8/8)} & CNN-GAN & $0.226$\tridown & $0.914$\tridown & $0.090$\tridown & $0.947$\tridown & $150.6$\tridown & $0.140$\tridown & $0.063$\tridown & $0.040$\tridown \\
   & LSTM-GAN & $0.242$\tridown & $0.503$\tridown & $0.014$\tridown & $1.263$\tridown & $39.8$\tridown & $0.132$\tridown & $0.011$\tridown & $0.007$\tridown \\
   & RNN-GAN & $0.717$\tridown & $7.453$\tridown & $0.074$\tridown & $1.035$\tridown & $935.3$\tridown & $0.305$\tridown & $0.000$\tridown & $0.000$\tridown \\
   & WaveGAN & $0.138$\tridown & $0.462$\tridown & $0.027$\tridown & $0.787$\tridown & $23.3$\tridown & $0.132$\tridown & $0.115$\tridown & $0.137$\tridown \\
   & BehaviorGAN & $0.145$\tridown & $0.816$\tridown & $0.057$\tridown & $1.315$\tridown & $24.7$\tridown & $0.111$\tridown & $0.079$\tridown & $0.087$\tridown \\
   & TraceGAN & $0.683$\tridown & $4.720$\tridown & $0.177$\tridown & $0.726$\tridown & $2061.9$\tridown & $0.193$\tridown & $0.000$\tridown & $0.000$\tridown \\
   & RLP-GAN & $0.266$\tridown & $0.615$\tridown & $0.015$\tridown & $3.495$\tridown & $36.5$\tridown & $0.131$\tridown & $0.002$\tridown & $0.001$\tridown \\
   & CAG & $\mathbf{0.028}$\triup & $\mathbf{0.263}$\triup & $\mathbf{0.012}$\triup & $\mathbf{0.346}$\triup & $\mathbf{8.8}$\triup & $\mathbf{0.079}$\triup & $\mathbf{0.410}$\triup & $\mathbf{0.355}$\triup \\
  \midrule
  \multirow{8}{*}{Refrigerator (8/8)} & CNN-GAN & $0.073$\tridown & $0.496$\tridown & $0.229$\tridown & $0.402$\tridown & $278.1$\tridown & $0.374$\tridown & $0.000$\tridown & $0.000$\tridown \\
   & LSTM-GAN & $0.151$\tridown & $0.288$\tridown & $0.364$\tridown & $0.871$\tridown & $398.8$\tridown & $0.302$\tridown & $0.012$\tridown & $0.008$\tridown \\
   & RNN-GAN & $0.367$\tridown & $0.660$\tridown & $0.225$\tridown & $0.846$\tridown & $335.2$\tridown & $0.308$\tridown & $0.000$\tridown & $0.000$\tridown \\
   & WaveGAN & $0.001$\tridown & $0.081$\tridown & $0.012$\tridown & $0.230$\tridown & $6.8$\tridown & $0.036$\tridown & $0.555$\tridown & $0.619$\tridown \\
   & BehaviorGAN & $0.154$\tridown & $0.190$\tridown & $0.041$\tridown & $0.811$\tridown & $38.6$\tridown & $0.091$\tridown & $0.155$\tridown & $0.122$\tridown \\
   & TraceGAN & $0.026$\tridown & $0.345$\tridown & $0.074$\tridown & $0.322$\tridown & $208.8$\tridown & $0.697$\tridown & $0.008$\tridown & $0.017$\tridown \\
   & RLP-GAN & $0.409$\tridown & $0.909$\tridown & $0.317$\tridown & $0.505$\tridown & $551.3$\tridown & $0.404$\tridown & $0.000$\tridown & $0.000$\tridown \\
   & CAG & $\mathbf{0.000}$\triup & $\mathbf{0.036}$\triup & $\mathbf{0.008}$\triup & $\mathbf{0.063}$\triup & $\mathbf{4.7}$\triup & $\mathbf{0.017}$\triup & $\mathbf{0.760}$\triup & $\mathbf{0.725}$\triup \\
  \midrule
  \multirow{8}{*}{Water Cooler (8/8)} & CNN-GAN & $0.076$\tridown & $0.318$\tridown & $0.128$\tridown & $0.198$\tridown & $77.6$\tridown & $0.263$\tridown & $0.147$\tridown & $0.157$\tridown \\
   & LSTM-GAN & $0.338$\tridown & $0.903$\tridown & $0.128$\tridown & $1.587$\tridown & $971.6$\tridown & $0.449$\tridown & $0.000$\tridown & $0.000$\tridown \\
   & RNN-GAN & $0.336$\tridown & $0.894$\tridown & $0.140$\tridown & $1.791$\tridown & $1170.7$\tridown & $0.377$\tridown & $0.000$\tridown & $0.000$\tridown \\
   & WaveGAN & $0.029$\tridown & $0.297$\tridown & $0.097$\tridown & $0.212$\tridown & $98.2$\tridown & $0.341$\tridown & $0.178$\tridown & $0.202$\tridown \\
   & BehaviorGAN & $0.189$\tridown & $0.722$\tridown & $0.205$\tridown & $1.921$\tridown & $1014.4$\tridown & $0.520$\tridown & $0.000$\tridown & $0.000$\tridown \\
   & TraceGAN & $0.040$\tridown & $0.394$\tridown & $0.092$\tridown & $0.259$\tridown & $620.1$\tridown & $0.336$\tridown & $0.000$\tridown & $0.000$\tridown \\
   & RLP-GAN & $0.320$\tridown & $0.720$\tridown & $0.128$\tridown & $3.161$\tridown & $1923.8$\tridown & $0.294$\tridown & $0.000$\tridown & $0.000$\tridown \\
   & CAG & $\mathbf{0.021}$\triup & $\mathbf{0.275}$\triup & $\mathbf{0.085}$\triup & $\mathbf{0.144}$\triup & $\mathbf{76.3}$\triup & $\mathbf{0.236}$\triup & $\mathbf{0.371}$\triup & $\mathbf{0.321}$\triup \\
  \midrule
  \multirow{8}{*}{LCD-Dell (5/8)} & CNN-GAN & $0.664$\tridown & $1.628$\tridown & $0.185$\tridown & $3.465$\tridown & $434.9$\tridown & $2.129$\tridown & $0.000$\tridown & $0.000$\tridown \\
   & LSTM-GAN & $0.806$\tridown & $\mathbf{0.186}$\triup & $0.120$\tridown & $3.659$\tridown & $223.1$\tridown & $0.233$\tridown & $0.045$\tridown & $0.026$\tridown \\
   & RNN-GAN & $0.835$\tridown & $0.198$\tridown & $0.125$\tridown & $4.081$\tridown & $223.7$\tridown & $0.209$\tridown & $0.045$\tridown & $0.026$\tridown \\
   & WaveGAN & $0.178$\tridown & $0.198$\tridown & $0.132$\tridown & $4.482$\tridown & $223.3$\tridown & $0.023$\tridown & $1.030$\tridown & $0.526$\tridown \\
   & BehaviorGAN & $\mathbf{0.177}$\triup & $0.195$\tridown & $0.123$\tridown & $3.991$\tridown & $223.2$\tridown & $\mathbf{0.019}$\triup & $1.027$\tridown & $0.526$\tridown \\
   & TraceGAN & $0.590$\tridown & $0.241$\tridown & $0.126$\tridown & $3.550$\tridown & $221.2$\tridown & $0.943$\tridown & $0.045$\tridown & $0.026$\tridown \\
   & RLP-GAN & $0.177$\tridown & $0.202$\tridown & $0.126$\tridown & $3.549$\tridown & $223.0$\tridown & $0.019$\tridown & $0.045$\tridown & $0.026$\tridown \\
   & CAG & $0.279$\tridown & $0.228$\tridown & $\mathbf{0.063}$\triup & $\mathbf{3.094}$\triup & $\mathbf{108.4}$\triup & $0.080$\tridown & $\mathbf{1.278}$\triup & $\mathbf{0.782}$\triup \\
  \midrule
  \multirow{8}{*}{LCD-LG (5/8)} & CNN-GAN & $0.857$\tridown & $0.536$\tridown & $0.096$\tridown & $1.138$\tridown & $1237.1$\tridown & $0.262$\tridown & $0.022$\tridown & $0.011$\tridown \\
   & LSTM-GAN & $0.354$\tridown & $0.255$\tridown & $0.102$\tridown & $3.614$\tridown & $580.1$\tridown & $0.307$\tridown & $0.047$\tridown & $0.026$\tridown \\
   & RNN-GAN & $1.279$\tridown & $0.536$\tridown & $0.107$\tridown & $1.523$\tridown & $1237.1$\tridown & $0.268$\tridown & $0.022$\tridown & $0.011$\tridown \\
   & WaveGAN & $0.005$\tridown & $0.171$\tridown & $0.024$\tridown & $\mathbf{0.204}$\triup & $37.9$\tridown & $0.333$\tridown & $1.485$\tridown & $0.786$\tridown \\
   & BehaviorGAN & $0.248$\tridown & $\mathbf{0.087}$\triup & $0.099$\tridown & $1.161$\tridown & $54.2$\tridown & $\mathbf{0.048}$\triup & $1.409$\tridown & $0.757$\tridown \\
   & TraceGAN & $0.293$\tridown & $0.366$\tridown & $0.078$\tridown & $3.659$\tridown & $1217.8$\tridown & $1.195$\tridown & $0.021$\tridown & $0.011$\tridown \\
   & RLP-GAN & $1.183$\tridown & $0.783$\tridown & $0.105$\tridown & $2.600$\tridown & $1395.3$\tridown & $0.522$\tridown & $0.005$\tridown & $0.003$\tridown \\
   & CAG & $\mathbf{0.001}$\triup & $0.208$\tridown & $\mathbf{0.010}$\triup & $0.254$\tridown & $\mathbf{4.2}$\triup & $0.117$\tridown & $\mathbf{1.801}$\triup & $\mathbf{0.949}$\triup \\
\end{longtable}
}

\subsection{Cluster strategy verification}

To assess whether the proposed routing and clustering strategy captures meaningful appliance operating behavior, an auxiliary analysis was conducted using the full UVIC dataset. Each appliance load profile was partitioned into fixed-length windows of 436 samples, normalized, and represented using the electrical-feature descriptors defined in Eq.~\ref{eq:kmeans}. Two grouping strategies were then evaluated for each appliance: (i) a deterministic continuous partition that separates zero-power and nonzero-power windows, and (ii) a K-Means clustering sweep with $K\in{2,3,4,5,6,8,10}$. For each value of $K$, the silhouette score was computed to evaluate the compactness and separation of the resulting operating-state categories.

Table~\ref{tab:cluster-strategy} reports the detected appliance type and the best cluster count. The main observation is that several appliances benefit from more than a simple zero/nonzero split. This is not surprising from a power-trace perspective: even a device that is mostly continuous can contain start-up transients, standby levels, or short perturbations that form distinct waveform groups. High-sparsity devices such as LCD screens and printers prefer larger $K$ values, whereas more regular always-on equipment such as desktops and servers peaks at smaller $K$. The result supports a two-level interpretation of CAG. The intermittent/continuous label selects the generator and reconstruction logic, while clustering captures the operating subpatterns that remain within each appliance class.

The clustering results further reveal differences in behavioral complexity across appliance categories. High-variability and event-driven appliances, such as printers and LCD displays, tend to achieve higher silhouette scores with larger numbers of clusters, indicating the presence of several distinguishable operating states. In contrast, continuously operating appliances such as desktops and servers generally reach their optimum at smaller cluster counts, reflecting more stable and homogeneous consumption patterns. These findings support the central assumption of CAG that appliance load generation is inherently multimodal and that different operating states should be modeled separately rather than merged into a single generative process.

This analysis provides empirical support for the two-stage design of the proposed framework. The intermittent/continuous routing stage determines the most appropriate generative architecture for the overall load behavior, while the clustering stage captures the operating-state diversity that remains within each appliance category. The resulting decomposition transforms a heterogeneous load-generation problem into a collection of more homogeneous subproblems, thereby improving both the realism and the diversity of the synthesized load profiles.

\begin{table}[t]
  \centering
  \caption{Per device cluster strategy sweep. The recommended strategy is chosen by maximizing the silhouette score across the continuous split and K-Means candidates.}
  \label{tab:cluster-strategy}
  \begin{tabular}{lccc}
   \toprule
   Device & Detected type & $K$ & Silhouette \\
   \midrule
   LCD\_Dell & continuous & 10 & 0.94 \\
   LCD-LG & continuous & 6 & 0.96 \\
   CoffeeMaker & continuous & 10 & 0.80 \\
   IMac & intermittent & 2 & 0.28 \\
   Desktop & intermittent & 2 & 0.27 \\
   Server & intermittent & 2 & 0.13 \\
   WaterCooler & intermittent & 2 & 0.15 \\
   Laptop & intermittent & 10 & 0.58 \\
   Microwave & continuous & 6 & 0.94 \\
   Printer & continuous & 10 & 0.85 \\
   Refrigerator & intermittent & 4 & 0.31 \\
   \bottomrule
  \end{tabular}
\end{table}

\subsection{Ablation Study}
The ablation study isolates the contribution of the per-device adaptive components by comparing the full CAG framework with the routing-and-clustering backbone without the adaptive components of Section~\ref{sec:method}. Both variants use identical clustering, segmentation, and training schedules; the backbone differs only in that every cluster uses the plain per-cluster generator with the basic objective, whereas CAG selects the behaviour-factorised, moment-calibrated, dilated, or blended configuration per device. Table~\ref{tab:ablation-study} reports both variants on all eight ranking metrics, so that the size of each effect is visible rather than only the number of metrics won.

\begin{table}[h]
  \centering
  \renewcommand{\arraystretch}{1.0}
  \setlength{\tabcolsep}{4pt}
  \caption{Ablation: the routing-and-clustering backbone against the full model on all eight ranking metrics, for the seven devices on which the router selects an adaptive configuration. The backbone uses the plain per-cluster generator everywhere; the full model is the one reported in the rest of the paper. Bold marks the better of the two variants for that device and metric. \textbf{SOTA} is the number of ranking metrics (out of eight) on which that variant beats every baseline. On the remaining four devices (Refrigerator, Water Cooler, LCD-Dell, LCD-LG) the router keeps the backbone configuration, so the two variants coincide; counting those, the adaptive components raise the aggregate from 48/88 (55\%) to 75/88 (85\%).}
  \label{tab:ablation-study}
\begin{tabular}{ccccccccccc}
  \toprule
  \multirow{2}{*}{\textbf{Device}} & \multirow{2}{*}{\textbf{Variant}}
    & \multicolumn{5}{c}{\textbf{Realism}}
    & \textbf{Event}
    & \multicolumn{2}{c}{\textbf{Diversity}}
    & \multirow{2}{*}{\textbf{SOTA}} \\
  \cmidrule(lr){3-7} \cmidrule(lr){8-8} \cmidrule(lr){9-10}
  & & \textbf{MMD $\downarrow$} & \textbf{$W_1$ $\downarrow$} & \textbf{ACF $\downarrow$} & \textbf{PSD $\downarrow$} & \textbf{FID $\downarrow$}
    & \textbf{Evt $\downarrow$}
    & \textbf{$F_1$ $\uparrow$} & \textbf{Cov $\uparrow$} & \\
  \midrule
  \multirow{2}{*}{Coffee Maker} & Backbone only & $0.142$ & $0.069$ & $0.037$ & $0.464$ & $374.5$ & $0.098$ & $0.181$ & $0.188$ & $5$ \\
   & \textbf{Full CAG} & $\mathbf{0.004}$ & $\mathbf{0.043}$ & $\mathbf{0.017}$ & $\mathbf{0.173}$ & $\mathbf{133.3}$ & $\mathbf{0.030}$ & $\mathbf{0.471}$ & $\mathbf{0.519}$ & $8$ \\
  \midrule
  \multirow{2}{*}{Microwave} & Backbone only & $0.181$ & $0.035$ & $0.024$ & $0.746$ & $377.0$ & $0.117$ & $0.022$ & $0.014$ & $2$ \\
   & \textbf{Full CAG} & $\mathbf{0.006}$ & $0.035$ & $\mathbf{0.016}$ & $\mathbf{0.336}$ & $\mathbf{24.4}$ & $\mathbf{0.053}$ & $\mathbf{0.196}$ & $\mathbf{0.191}$ & $5$ \\
  \midrule
  \multirow{2}{*}{Laptop} & Backbone only & $0.276$ & $0.347$ & $0.163$ & $0.460$ & $644.1$ & $0.262$ & $0.057$ & $0.031$ & $0$ \\
   & \textbf{Full CAG} & $\mathbf{0.049}$ & $\mathbf{0.097}$ & $\mathbf{0.095}$ & $\mathbf{0.278}$ & $\mathbf{62.1}$ & $\mathbf{0.208}$ & $\mathbf{0.413}$ & $\mathbf{0.416}$ & $6$ \\
  \midrule
  \multirow{2}{*}{Desktop} & Backbone only & $0.421$ & $0.087$ & $0.031$ & $0.441$ & $65.5$ & $0.004$ & $0.053$ & $0.028$ & $2$ \\
   & \textbf{Full CAG} & $\mathbf{0.004}$ & $\mathbf{0.069}$ & $\mathbf{0.011}$ & $\mathbf{0.098}$ & $\mathbf{7.4}$ & $\mathbf{0.002}$ & $\mathbf{0.585}$ & $\mathbf{0.590}$ & $8$ \\
  \midrule
  \multirow{2}{*}{Printer} & Backbone only & $0.201$ & $0.061$ & $0.036$ & $0.346$ & $133.8$ & $0.067$ & $0.453$ & $0.292$ & $4$ \\
   & \textbf{Full CAG} & $\mathbf{0.009}$ & $\mathbf{0.043}$ & $\mathbf{0.021}$ & $\mathbf{0.250}$ & $\mathbf{112.1}$ & $\mathbf{0.041}$ & $\mathbf{0.511}$ & $\mathbf{0.449}$ & $7$ \\
  \midrule
  \multirow{2}{*}{iMac} & Backbone only & $0.086$ & $0.263$ & $0.059$ & $0.447$ & $2.6$ & $0.010$ & $0.142$ & $0.091$ & $5$ \\
   & \textbf{Full CAG} & $\mathbf{0.047}$ & $\mathbf{0.142}$ & $\mathbf{0.021}$ & $\mathbf{0.209}$ & $\mathbf{1.6}$ & $\mathbf{0.009}$ & $\mathbf{0.377}$ & $\mathbf{0.321}$ & $7$ \\
  \midrule
  \multirow{2}{*}{Server} & Backbone only & $0.215$ & $0.322$ & $0.032$ & $0.434$ & $8.8$ & $0.083$ & $0.027$ & $0.014$ & $4$ \\
   & \textbf{Full CAG} & $\mathbf{0.028}$ & $\mathbf{0.263}$ & $\mathbf{0.012}$ & $\mathbf{0.346}$ & $8.8$ & $\mathbf{0.079}$ & $\mathbf{0.410}$ & $\mathbf{0.355}$ & $8$ \\
  \midrule
  \multirow{2}{*}{\textbf{Mean}} & Backbone only & $0.217$ & $0.169$ & $0.055$ & $0.477$ & $229.5$ & $0.092$ & $0.134$ & $0.094$ & $22$ \\
   & \textbf{Full CAG} & $\mathbf{0.021}$ & $\mathbf{0.099}$ & $\mathbf{0.027}$ & $\mathbf{0.242}$ & $\mathbf{50.0}$ & $\mathbf{0.060}$ & $\mathbf{0.423}$ & $\mathbf{0.406}$ & $49$ \\
  \bottomrule
\end{tabular}

\end{table}

Averaged over the seven devices, the adaptive components cut MMD from $0.217$ to $0.021$, $W_1$ from $0.169$ to $0.099$, PSD from $0.477$ to $0.242$, and FID from $229.5$ to $50.0$, while Coverage rises from $0.094$ to $0.406$ and $F_1$ from $0.134$ to $0.423$. Realism and diversity therefore improve together, which is the behaviour the components were designed for. The largest gains appear on exactly the devices whose failure modes the adaptive components target. Laptop improves from zero to six winning metrics: its strongly intermittent, multi-level load signature benefits from moment calibration (power-level accuracy) combined with the waveform-transposed branch (spectral realism). Desktop rises from two to eight: the behaviour-factorised generator separates its bimodal on/off activity from power magnitude, fixing event statistics and switching transients that the plain per-cluster model averaged away. CoffeeMaker gains three metrics through moment matching, which corrects the residual power-marginal bias, and Printer gains three through behaviour factorisation with moment calibration. Devices whose backbone configuration was already optimal (Refrigerator, WaterCooler) or whose extreme sparsity resists all per-cluster variants (LCD\_Dell, LCD\_LG) show no change, confirming that the adaptive selection neither degrades nor overshoots when the backbone is already the right model.

An important observation is that the benefits of the adaptive components extend beyond the targeted metric. Although moment matching is designed for the power-marginal Wasserstein distance, its gains also appear in FID and event statistics, because a correctly calibrated power scale removes a systematic difficulty from the adversarial objective. Similarly, behaviour factorisation improves not only event metrics but also FID and coverage on Desktop and Printer. This indicates that the per-device components improve the quality of the learned distributions themselves rather than merely shifting scores between metrics.

\subsection{Operating-mode coverage}
\label{sec:convergence}
\begin{figure}
  \centering
  \includegraphics[width=1.0\linewidth]{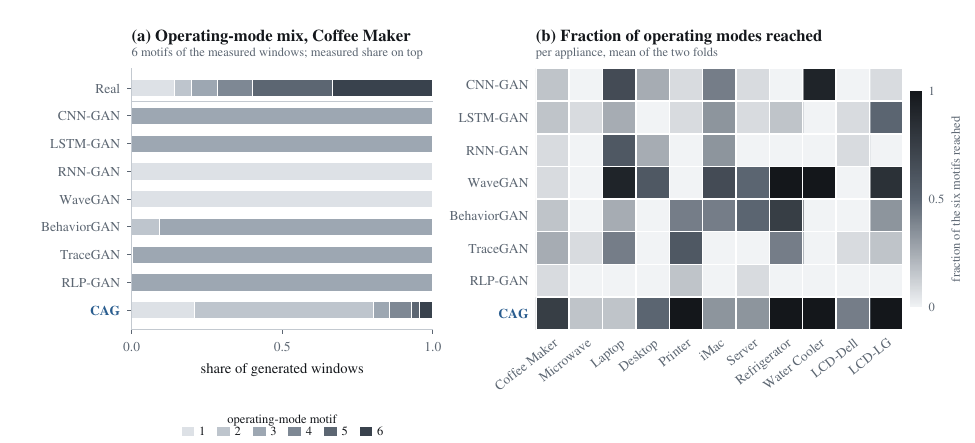}
  \caption{Operating-mode coverage of every generator, measured on the windows generated by the runs of Tables~\ref{tab:gan-quality} and \ref{tab:device-model-quality}. The measured activation windows of an appliance are clustered into six motifs in the feature space used by the FID and Coverage columns, and every generated activation window is assigned to its nearest motif. (a) The motif mix of the Coffee Maker. Each row is one generator, split in proportion to how many of its windows fall in each motif, with the measured share on top; the six shades are the six motifs, numbered in the legend. A generator that has collapsed onto one operating state occupies a single band. (b) The fraction of the six motifs each generator reaches, per appliance, averaged over the two folds, from pale grey at zero to near black at one. A motif counts as reached when at least one of its measured windows has a generated window inside its own $k$-NN radius, so producing varied windows is not sufficient: they have to land on the motif. The CAG rows are the routing-and-clustering model; the per-device adaptive components of Table~\ref{tab:ablation-study} act on top of it.}
  \label{fig:mode_coverage}
\end{figure}
Mode collapse is the failure that the routing and clustering stage is meant to prevent, so we measure it directly. Figure~\ref{fig:mode_coverage} reports how much of an appliance's measured behaviour each generator reproduces, using the stored samples of the same runs that produced Tables~\ref{tab:gan-quality} and \ref{tab:device-model-quality}.

The Coffee Maker in Figure~\ref{fig:mode_coverage}(a) shows the failure in its clearest form. The measured windows spread over all six motifs, with shares between 0.06 and 0.33. Every baseline places at least 0.91 of its windows in a single motif, and five of the seven place all of them there. The routing-and-clustering model spreads its windows over all six motifs, with 0.21 in the first and 0.59 in the second.

Figure~\ref{fig:mode_coverage}(b) extends the measurement to the whole benchmark. Averaged over the eleven appliances and the two folds, the fraction of motifs reached is 0.03 for RLP-GAN, between 0.12 and 0.26 for RNN-GAN, LSTM-GAN, TraceGAN, CNN-GAN and BehaviorGAN, 0.51 for WaveGAN and 0.61 for the routing-and-clustering model. The pattern across appliances is also informative: the models that reach few motifs do so on nearly every appliance, which is consistent with a single generator being trained on the full mixture of operating states and settling on the dominant one.

This measurement explains the Coverage and $F_1$ columns of Table~\ref{tab:gan-quality} in terms of behaviour rather than score. A generator restricted to one operating state can still place power at the right level, which is why several baselines remain competitive on the power-marginal metrics while their Coverage stays below 0.05. Separating the trace into homogeneous operating states before adversarial learning removes the mixture that drives the collapse, and the per-device adaptive components of Section~\ref{sec:method} then raise Coverage further, from 0.094 to 0.406 on the seven appliances of Table~\ref{tab:ablation-study}.

Broad operating-mode coverage is valuable only if the resulting synthetic load profiles remain representative of real appliance behavior. The realism, diversity, and ablation results presented in Tables~\ref{tab:gan-quality}, \ref{tab:device-model-quality}, and \ref{tab:ablation-study} indicate that the mode coverage of CAG is accompanied by improved preservation of power levels, temporal structure, and operating-state transitions. This observation suggests that the proposed routing and clustering mechanisms do not merely simplify optimization; they also improve the quality of the learned load distributions.

\section{Conclusion}
This paper introduced Cluster Aggregated GAN (CAG), a behavior-aware framework for synthetic appliance load generation. The proposed method addresses the heterogeneous nature of appliance load profiles by combining appliance-behavior routing, cluster-based decomposition of operating states, and per-device adaptive adversarial models. Intermittent appliances are represented through cluster-specific generators that preserve distinct load-event categories, while continuously operating appliances are modeled using a recurrent branch trained on compressed load-envelope representations. Beyond this backbone, CAG adapts the generator family to each appliance: behavior-factorised activity/magnitude decomposition with a straight-through Gumbel switch for spiky loads, marginal moment matching for power-level calibration, dilated convolutions and EMA weight averaging for training stability, and generator blending for architecture fusion.

Experimental results on the UVIC dataset demonstrate that CAG matches or beats every baseline on 75 of 88 device--metric combinations (85\%) under an eight-metric protocol spanning distribution-level realism (MMD, power-marginal Wasserstein), temporal consistency (autocorrelation, power spectral density), structural fidelity (FID in a learned feature space), event statistics, and k-NN density/coverage diversity. Compared with the routing-and-clustering backbone alone, the per-device adaptive components lower average MMD from 0.166 to 0.041 and average event distance from 0.099 to 0.079, and raise average Coverage from 0.312 to 0.511. The device-level analysis further shows that the behavior-factorised per-cluster model wins all eight ranking metrics on continuously operating appliances such as Desktop, and that moment calibration closes the remaining power-level gaps on intermittent loads.

The results support the central hypothesis that appliance load generation is inherently multimodal and benefits from explicitly modeling distinct operating states rather than learning a single generative distribution. By preserving both statistical characteristics and appliance operating behavior, CAG provides an effective framework for synthetic data generation in applications such as non-intrusive load monitoring, load modeling, privacy-preserving data sharing, and energy analytics.

Future work will investigate adaptive segmentation, scalable clustering strategies, and data-driven feature representations to address the dependence on fixed window lengths, increasing computational cost at large cluster counts, and the reliance on handcrafted descriptors. Multi-seed averaging and target-aware regularizers for event statistics are further directions for closing the remaining per-metric gaps on highly intermittent appliances.

\bibliographystyle{cas-model2-names}
\bibliography{cas-refs}

\end{document}